\documentclass{article}

    \PassOptionsToPackage{numbers, compress}{natbib}

    \usepackage[preprint]{neurips_2023}

\usepackage{amssymb} 
\usepackage{multirow}
\usepackage[utf8]{inputenc} 
\usepackage[T1]{fontenc}    
\usepackage{url}            
\usepackage{booktabs}       
\usepackage{amsfonts}       
\usepackage{nicefrac}       
\usepackage{microtype}      
\usepackage{xcolor}         
\usepackage{graphicx}
\definecolor{citecolor}{HTML}{0071BC}
\definecolor{linkcolor}{HTML}{ED1C24}
\definecolor{highlightcolor}{HTML}{ABCDEF}
\usepackage[pagebackref, breaklinks, colorlinks, letterpaper=true, citecolor=citecolor, linkcolor=linkcolor, bookmarks=false]{hyperref}

\newtheorem{proof}{Proof}[section]

\usepackage{amsmath,amsfonts,bm,amssymb}
\usepackage[most]{tcolorbox}

\def\1{\bm{1}}

\def\vmu{{\bm{\mu}}}
\def\vtheta{{\bm{\theta}}}

\def\vepsilon{{\bm{\epsilon}}}

\def\vc{{\bm{c}}}

\def\ve{{\bm{e}}}

\def\vv{{\bm{v}}}

\def\vx{{\bm{x}}}

\def\vz{{\bm{z}}}

\def\mI{{\bm{I}}}

\DeclareMathAlphabet{\mathsfit}{\encodingdefault}{\sfdefault}{m}{sl}
\SetMathAlphabet{\mathsfit}{bold}{\encodingdefault}{\sfdefault}{bx}{n}

\def\gD{{\mathcal{D}}}
\def\gE{{\mathcal{E}}}

\def\gI{{\mathcal{I}}}

\def\gL{{\mathcal{L}}}

\def\gN{{\mathcal{N}}}

\def\gV{{\mathcal{V}}}

\newcommand{\E}{\mathbb{E}}

\usepackage[most]{tcolorbox}

\newtcolorbox{mblock}[1]
{
  colframe     = gray,
  coltitle     = black,
  colbacktitle = lightgray!50!white,
  colback      = white,
  title        = #1,
  fonttitle    = \bfseries,
  arc          = 0mm,
  left         = 2pt,
  right        = 2pt,
}

\newtcolorbox{rblock}[1]
{
  colframe     = green,
  coltitle     = gray,
  colbacktitle = lightgray!50!white,
  colback      = white,
  title        = \ifthenelse{\isempty{#1}}
                  {Remark}
                  {Remark: #1},
  fonttitle    = \bfseries,
  arc          = 0mm,
  left         = 2pt,
  right        = 2pt,
}

\def\ie{\emph{i.e.,}}
\def\eg{\emph{e.g.,}}

\usepackage{etoc}
\etocdepthtag.toc{mtchapter}
\etocsettagdepth{mtchapter}{subsection}
\etocsettagdepth{mtappendix}{none}

\makeatletter
\renewcommand\@fnsymbol[1]{\ensuremath{^\dagger}}
\makeatother

\title{Gen-L-Video: Multi-Text to Long Video Generation via Temporal Co-Denoising}

\author{Fu-Yun Wang\textsuperscript{4,1,2} \hspace{-3mm} \quad Wenshuo Chen\textsuperscript{5} \hspace{-3mm} \quad Guanglu Song\textsuperscript{3} \hspace{-3mm}\quad Han-Jia Ye\textsuperscript{4} \hspace{-3mm} \quad Yu Liu\textsuperscript{1}\thanks{Correspondence to: Hongsheng Li~(hsli@ee.cuhk.edu.hk), Yu Liu~(liuyuisanai@gmail.com)}\hspace{-2mm} \quad Hongsheng Li\textsuperscript{2}\footnotemark[2]\\
\textsuperscript{1}Shanghai AI Laboratory
\quad \textsuperscript{2}Multimedia Laboratory, The Chinese University of Hong Kong \\
\textsuperscript{3}Sensetime Research 
\quad \textsuperscript{4}Nanjing University
\quad \textsuperscript{5}Tsinghua University 
\\
\texttt{\{wangfuyun@smail, yehj@lamda\}.nju.edu.cn}\quad \texttt{cws21@mails.tsinghua.edu.cn}\\
\texttt{songguanglu@sensetime.com} \quad \texttt{liuyuisanai@gmail.com} \quad \texttt{hsli@ee.cuhk.edu.hk}
}
\begin{document}

\maketitle

\begin{abstract}
Leveraging large-scale image-text datasets and advancements in diffusion models, text-driven generative models have made remarkable strides in the field of image generation and editing. This study explores the potential of extending the text-driven ability to the generation and editing of multi-text conditioned long videos. Current methodologies for video generation and editing, while innovative, are often confined to extremely short videos (typically less than 24 frames) and are limited to a single text condition. These constraints significantly limit their applications given that real-world videos usually consist of multiple segments, each bearing different semantic information. To address this challenge, we introduce a novel paradigm dubbed as \textit{Gen-L-Video} capable of extending off-the-shelf short video diffusion models for generating and editing videos comprising hundreds of frames with diverse semantic segments without introducing additional training, all while preserving content consistency. We have implemented three mainstream text-driven video generation and editing methodologies and extended them to accommodate longer videos imbued with a variety of semantic segments with our proposed paradigm. Our experimental outcomes reveal that our approach significantly broadens the generative and editing capabilities of video diffusion models, offering new possibilities for future research and applications. Code is available at: \url{https://github.com/G-U-N/Gen-L-Video}.
\end{abstract}

\vspace{-1em}
\section{Introduction}
\vspace{-0.4em}
Benefitting from pre-training on large-scale text-image datasets~\cite{schuhmann2022laion} and the development and refinement of the diffusion model~\cite{nichol2021glide, ramesh2022hierarchical,ding2022cogview2, singer2022make, saharia2022photorealistic}, we have witnessed a plethora of successful applications, including impressive image generation, editing, and even fine-grained generation control through the injection of additional layout information~\cite{mou2023t2i,zhang2023adding}. A logical progression of this approach is its extension to the video realm for text-driven video generation and editing~\cite{singer2022make,qi2023fatezero,ho2022video,wu2022tune}.

Currently, there are three primary strategies for text-driven video generation and editing:
\begin{itemize}
    \itemsep0em 
    \item \textbf{Pretrained Text-to-Video (pretrained t2v)}~\cite{van2017neural,wu2022nuwa,hong2022cogvideo} involves training the diffusion model on a large-scale text-video paired dataset such as WebVid-10M~\cite{bain2021frozen}. Typically, a temporal interaction module, like Temporal Attention~\cite{ho2022video}, is added to the denoising model, fostering inter-frame information interaction to ensure frame consistency.
    \item \textbf{Tuning-free Text-to-Video (tuning-free t2v)}~\cite{qi2023fatezero,liu2023video,ceylan2023pix2video} utilizes the pre-trained Text-to-Image model to generate and edit video frame-by-frame, while applying additional controls to maintain consistency across frames (for instance, copying and modifying the attention map~\cite{hertz2022prompt,qi2023fatezero,liu2023video}, sparse causal attention~\cite{wu2022tune}, etc.).
    \item \textbf{One-shot tuning Text-to-Video (one-shot-tuning t2v)}~\cite{wu2022tune,nikankin2022sinfusion} proposes to fine-tune the pretrained text-to-image generation model on a single video instance to generate videos with similar motions or contents. Despite the extra training cost, one-shot tuning-based methods often offer more editing flexibility compared to tuning-free based methods. As depicted in Fig.~\ref{fig:comp}, both attempt to substitute the rabbit in the source video with a tiger or a puppy. The outcome produced by the tuning-free t2v method reveals elongated ears, losing authenticity. One-shot-tuning-based method, in contrast, effectively circumvents this problem.
\end{itemize}

Despite significant advances made by previous methods, they are accompanied by some fatal limitations, restricting their practical applications. First of all, the number of video frames generated by these methods is usually limited, generally less than 24 frames~\cite{qi2023fatezero,wu2022tune,liu2023video,khachatryan2023text2video,luo2023videofusion}. On one hand, the computational complexity of temporal attention scales quadratically with the number of frames, making the direct generation of ultra-long videos infeasible. On the other hand, ensuring consistency becomes more challenging with the increase in the number of frames. Another noteworthy limitation is that these methods typically generate videos controlled by a single text condition and cannot accommodate multiple text prompts. In reality, the content of a video often changes over time, meaning that a comprehensive video often comprises multiple segments each bearing different semantic information. This necessitates the development of video generation methods that can handle multiple text conditions. Though there are already attempts at generating long videos, they typically require additional training on large-scale text-video datasets and follow the autoregressive mechanism ~(\ie the generation of later frames is conditioned on former ones), which suffers from severe content degradation and inference inefficiency~(see Sec.~\ref{sec:long video generation} for more discussion).

\begin{figure*}[t!]
\vspace{-2em}
    \centering
    \includegraphics[width=\linewidth]{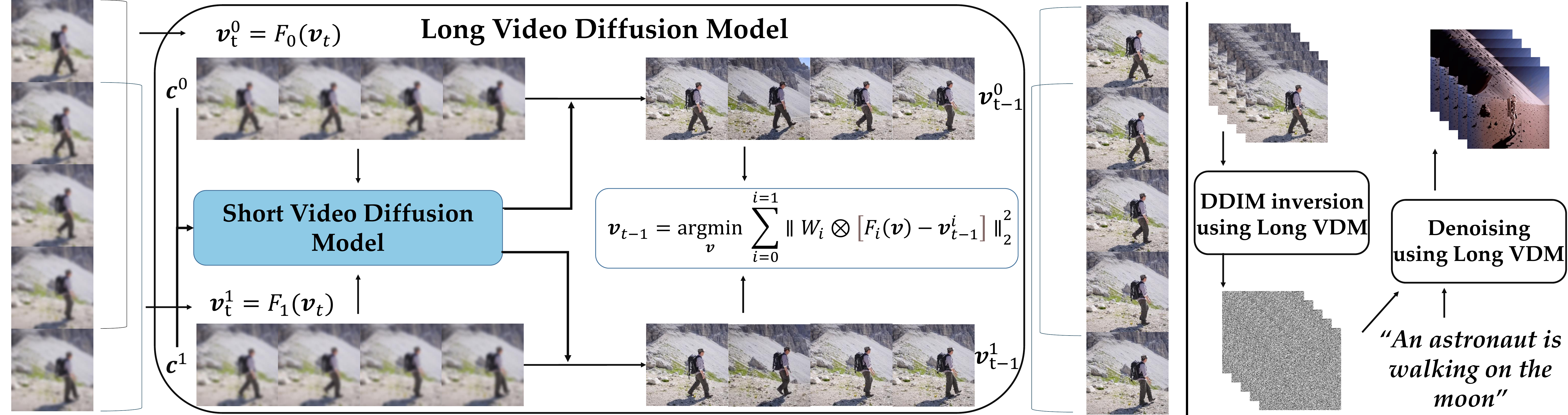}
    \caption{\textit{\textbf{Left: High-level overview of Temporal Co-Denoising.}} Our framework treats long videos of arbitrary lengths and multiple semantic segments as a collection of short videos with temporal overlapping. It allows us to effectively apply short video diffusion models to approximate the denoising path of these extended videos, ensuring consistency and coherence throughout. \textit{\textbf{Right: The pipeline of our framework for video editing.}} With the extended long video diffusion model, we first invert the given long video into an approximated initial noise through DDIM inversion. Then, we sample a new video guided by the given prompts~(either single or multiple). }
    \label{fig:lvdm}
    \vspace{-1em}
\end{figure*}

In light of these challenges, we propose a novel framework aimed at generating long videos with consistent, coherent content across multiple semantic segments. Unlike previous methods, we do not construct or train a long-video generator directly. Instead, we view a video as a collection of short video clips, each possessing independent semantic information. Hence, a natural idea is that generation of long videos can be seen as the direct splicing of multiple short videos. However, this simplistic division falls short of generating long videos with consistent content, resulting in noticeable content and detail discrepancies between different video clips. As shown in the third row of Fig.~\ref{fig:comp_co}, the color of the jeep car changes drastically among different video clips when they are denoised isolatedly. To counter this, we perceive long videos as short video clips with \textit{temporal overlapping}. We demonstrate that under certain conditions, the denoising path of a long video can be approximated by joint denoising of overlapping short videos in the temporal domain. In particular, as depicted in Fig.~\ref{fig:lvdm}, the noisy long video is initially mapped into multiple noisy short video clips via a designated function. Subsequently, existing off-the-shelf short video diffusion models can be employed to denoise these video clips under the guidance of various text conditions. These denoised short videos are then merged and inverted back to a less noisy original long video. Essentially, this procedure establishes an abstract long video generator and editor without necessitating any additional training, enabling the generation and editing of videos of any length using established short video generation and editing methodologies.

Our method was tested in three scenarios: pretrained t2v, tuning-free t2v, and one-shot-tuning t2v, all of which yielded favorable results. Furthermore, the incorporation of additional control information and advanced open-set detection~\cite{liu2023grounding} and segmentation~\cite{kirillov2023segment} technologies allows for more impressive results, such as precise layout control and arbitrary object video inpainting. Extensive experimental results validate the broad applicability and effectiveness of our proposed Gen-L-Video.

\section{Related Work}\label{sec:related work}

As we mentioned, the current mainstream strategies for video generation and editing can be mainly categorized into three types: pretrained Text-to-Video~(pretrained t2v), tuning-free Text-to-Video~(tuning-free t2v), and one-shot-tuning Text-to-Video~(one-shot-tuning t2v). The breakthroughs in video generation and editing techniques have largely drawn from the existing technologies for image editing and generation. In this section, we first introduce the development of text-to-image technology and then provide a brief overview of the key accomplishments of each of the three strategies. In the end, we discuss recent advances in long video generations and the advantage of Gen-L-Video over them.

\paragraph{Text-to-Image generation.}
Many early works~\cite{zhu2019dm,zhang2021cross,ye2021improving,xu2018attngan,tao2020df} train GANs~\cite{goodfellow2020generative} on image captioning datasets to produce text-conditional image samples. Other works~\cite{ramesh2021zero, yu2022scaling, yu2021vector, ding2022cogview2, gafni2022make} apply vector quantization~\cite{van2017neural} and then adopt autoregressive transformers to predict image tokens followed by text tokens.  Recently, several works~\cite{nichol2021glide, saharia2022photorealistic, gu2022vector, rombach2022high} adopt diffusion~\cite{ho2020denoising} models for Text-to-Image Generation. GLIDE~\cite{nichol2021glide} introduces classifier-free guidance~\cite{ho2022classifier} in the diffusion model to enhance image quality, while DALLE-2~\cite{ramesh2022hierarchical} improves text-image alignment using the CLIP~\cite{radford2021learning} feature space. Imagen~\cite{saharia2022photorealistic} employs cascaded diffusion models for high-definition video generation. VQ-diffusion~\cite{gu2022vector} and Latent Diffusion Model (LDM, also known as Stable Diffusion)~\cite{rombach2022high} train diffusion models in an autoencoder's latent space to boost efficiency. Variants of LDM are fine-tuned to achieve more functionality like inpainting, image variants, etc. ControlNet~\cite{zhang2023adding}, T2I-adapter~\cite{mou2023t2i} add new modules to accept additional image inputs, achieving precise generative layout control. Many fine-tuning strategies~\cite{ruiz2022dreambooth,hu2021lora,gal2022image} are also developed to force diffusion models to generate new concepts and styles, which shares a similar idea to continual learning~\cite{zhou2021pycil,wang2022foster,smith2023continual}.
\paragraph{Pretrained Text-to-Video.}
Despite significant advancements in Text-to-Image generation, generating videos from text remains a challenge due to the scarcity of high-quality, large-scale text-video datasets and the inherent complexity of modeling temporal consistency and coherence. Early works~\cite{mittal2017sync, pan2017create, marwah2017attentive, li2018video, gupta2018imagine, liu2019cross} primarily focus on generating videos in simple domains, such as moving digits or specific human actions. GODIVA~\cite{van2017neural} is the first model to utilize VQ-VAE and sparse attention for Text-to-Video generation, enabling more realistic scenes. NÜWA~\cite{wu2022nuwa} builds upon GODIVA with a unified representation for various generation tasks via multitask learning. CogVideo~\cite{hong2022cogvideo} incorporates additional temporal attention modules on top of the pre-trained Text-to-Image model~\cite{ding2022cogview2}. Similarly, Video Diffusion Models (VDM)~\cite{ho2022video} use a space-time factorized U-Net with joint image and video data training. Imagen Video~\cite{ho2022imagen} improves VDM with cascaded diffusion models and v-prediction parameterization for high-definition video generation. Make-A-Video~\cite{singer2022make}, MagicVideo~\cite{zhou2022magicvideo}, and LVDM~\cite{he2022latent} share similar motivations, aiming to transfer progress from t2i to t2v generation. Video Fusion~\cite{luo2023videofusion} decomposes the denoising process by resolving per-frame noise into base noise and residual noise to reflect the connections among frames. For video editing, Dreamix~\cite{molad2023dreamix} and Gen-1~\cite{esser2023structure} utilize the video diffusion model for video editing. 
\paragraph{Tuning-free Text-to-Video.}
It’s nontrivial to directly apply pretrained Text-to-Image model for video generation or editing without tuning. Recent diffusion-based image editing models~\cite{meng2021sdedit, hertz2022prompt, couairon2022diffedit, wu2022unifying, kawar2022imagic, tumanyan2022plug}, although powerful in processing individual frames in a video, results in inconsistencies between frames due to the models' lack of temporal awareness. Tune-A-Video~\cite{wu2022tune} finds that extending spatial self-attention to sparse causal attention with pretrained weight produces consistent content across frames. Fate-Zero~\cite{qi2023fatezero} and Video-P2P~\cite{liu2023video} apply sparse causal attention and attention control proposed in prompt2prompt, achieving consistent video editing. Pix2Video~\cite{ceylan2023pix2video} adds additional regularization to penalize the dramatic frame changes. Text2Video-Zero~\cite{khachatryan2023text2video} first proposes to generate videos in zero-shot settings with only pretrained text-to-image diffusion model. It applies the sparse causal attention and object mask to preserve the content consistency among frames and add motion dynamics in different scales to enrich the base latent code to generate consecutive motions. 
\paragraph{One-shot tuning Text-to-Video.}
Single-video GANs~\cite{arora2021singan-gif,gur2020hierarchical} generate new videos with similar appearance and dynamics to the input video, while they suffer from extensive computational burden. SinFusion~\cite{nikankin2022sinfusion} adapts diffusion models to single-video tasks and enables autoregressive video generation with improved motion generalization capabilities. Tune-A-Video~\cite{wu2022tune} proposes to fine-tune the pretrained text-to-image diffusion model on a video to enable generation of videos with similar motions, demonstrating powerful editing ability. 
\paragraph{Long video generation.}\label{sec:long video generation}
The generation of long videos has garnered significant attention in recent years, resulting in various attempts to address the challenges associated with this task~\cite{voleti2022masked,yu2023video,harvey2022flexible,ge2022long,liang2022nuwa}. Existing approaches typically rely on autoregressive models, such as NUWA-Infinity~\cite{liang2022nuwa}, Phenaki~\cite{villegas2022phenaki}, and TATS~\cite{ge2022long}, or diffusion models, including MCVD~\cite{voleti2022masked}, FDM~\cite{harvey2022flexible}, PVDM~\cite{yu2023video}, and LVDM~\cite{he2022latent}. All these methods employ an autoregressive mechanism for long video generation, wherein the generated frames serve as conditioning for subsequent frames. However, this mechanism often leads to significant content degradation after several extrapolations due to error accumulation. Furthermore, the autoregressive mechanism constrains generation to a sequential process, substantially reducing efficiency. Recently, NUWA-XL~\cite{yin2023nuwa} proposed a novel hierarchical diffusion process that enables parallel long video generation. Despite its advantages, this approach necessitates extensive pretraining on large long video datasets and requires a well-designed global diffusion model to generate key frames at the outset.  Our framework, instead of directly training or constructing a long video diffusion model, can approximate the arbitrary length long video denoising path with parallel joint denoising of off-the-shelf short video generation or editing models. In general, Gen-L-Video presents an efficient, convenient, and scalable paradigm for long video generation, addressing the limitations of existing methods and offering new possibilities for future research and applications. We make a direct comparison of our method with existing methods in Table~\ref{tab:comparison box}, and our approach demonstrates significant superiority.
\section{Method}
\begin{figure*}[t!]
\vspace{-3em}
    \centering
    \includegraphics[width=\linewidth]{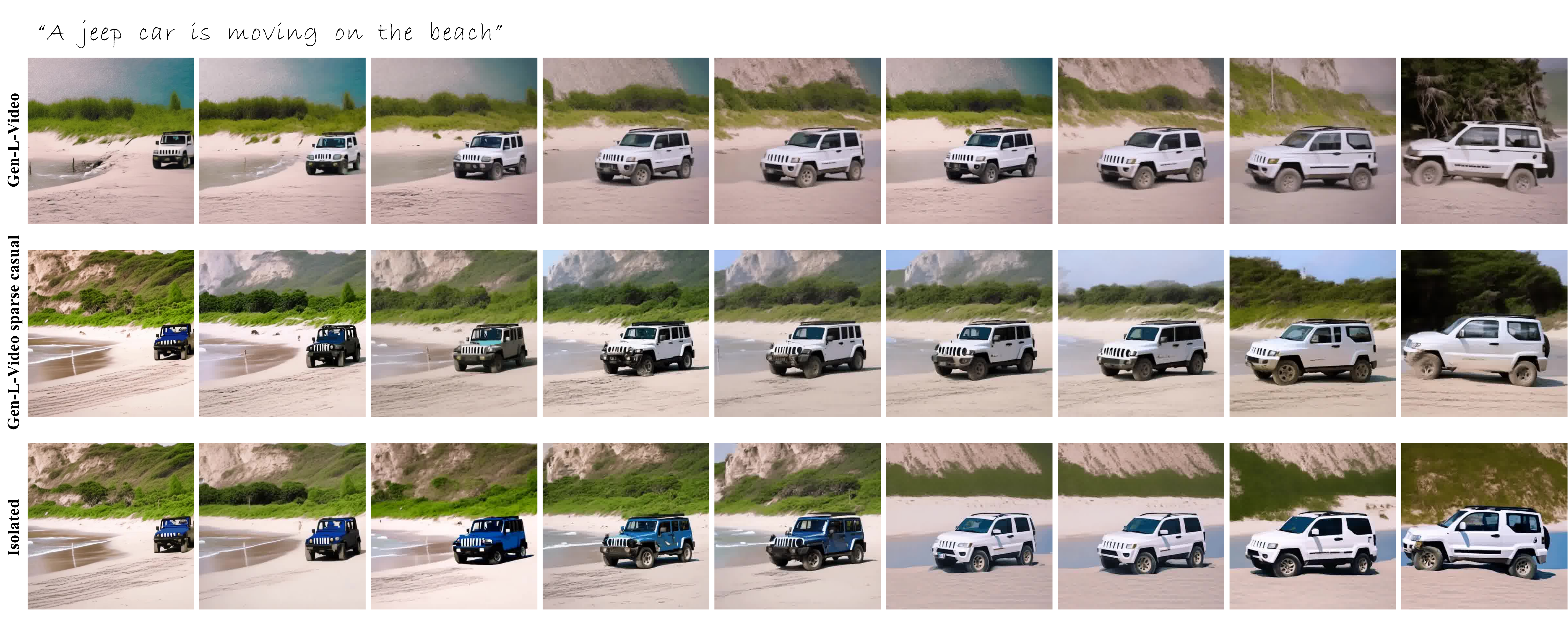}
    \caption{\textit{\textbf{Comparison between our method Gen-L-Video and isolated denoising with short video diffusion models.}} The long video consist of two short video clips~(each with 20 frames), and we sample one in five for better visualization. Gen-L-Video achieves the most smooth and consistent generation. Gen-L-Video sparse causal means we replace the bi-directional cross-frame attention with the sparse causal attention proposed in \cite{wu2022tune}, which typically leads to the first few frames inconsistent with the later ones, and we analyze the subtle reason in Sec.~\ref{sec:application}.}
    \label{fig:comp_co}
    \vspace{-1em}
\end{figure*}
\subsection{Preliminaries}
\textbf{Diffusion models}~\cite{ho2020denoising} perturb the data by gradually injecting noise to data $\vx_0 \sim q(\vx_0)$, which is formalized by a Markov chain:
\begin{align*}
    & q(\vx_{1:T}|\vx_0) = \prod\limits_{t=1}^T q(\vx_t|\vx_{t-1}),
    & q(\vx_t|\vx_{t-1}) = \gN(\vx_t|\sqrt{\alpha_t} \vx_{t-1}, \beta_t \mI),
\end{align*}
where $\beta_t$ is the noise schedule and $\alpha_t = 1 - \beta_t$. The data can be generated by reversing this process, \ie we gradually denoise to restore the original data. The diffusion model $p_\vtheta(\vx_{t-1}|\vx_t)$ parameterized by $\vtheta$ is trained to approximate the reverse transition $q(\vx_{t-1}|\vx_{t},\vx_0)$, which is formulated as 
\begin{align*}
        q(\vx_{t-1}|\vx_t, \vx_0) & = \gN(\vx_{t-1}; \Tilde{\vmu}_t(\vx_t, \vx_0), \Tilde{\beta}_t \mI), & \Tilde{\vmu}_t\left(\vx_t, \vx_0\right)=\frac{1}{\sqrt{\alpha_t}} \vx_t-\frac{1-\alpha_t}{\sqrt{1-\bar{\alpha}_t} \sqrt{\alpha_t}} \vepsilon,
\end{align*}
where $ \Bar{\alpha}_t = \prod_{s=1}^t \alpha_s$, $\Tilde{\beta}_t = \frac{1 - \Bar{\alpha}_{t-1}}{1 - \Bar{\alpha}_t} \beta_t$, and $\vepsilon$ is the noise injected to $\vx_0$ to obtain $\vx_t$. Therefore, the learning of $p_\vtheta(\vx_{t-1}|\vx_t)$ is equivalent to the learning a noise prediction network $\vepsilon_{\vtheta}(\vx_t,{t})$:
\begin{align*}
    \min\limits_\vtheta \E_{t, \vx_0, \vepsilon} \| \vepsilon - \vepsilon_\vtheta(\vx_t, t) \|_2^2, 
\end{align*}
where $t$ is uniformly sampled from $\{1,2,\dots,T\}$ and $\vx_t = \sqrt{\overline{\alpha}_t} \vx_0 + \sqrt{1 - \overline{\alpha}_t} \vepsilon$.

\textbf{DDIM}~\cite{song2020denoising} generalize the framework of DDPM and propose a deterministic ODE process, achieving faster sampling speed. The inversion trick of DDIM~\cite{mokady2022null}, based on the assumption that the ODE process can be reversed in the limit of small steps, can be used to approximate the corresponding noise of the given instance:
\begin{align*}
    \frac{\vx_{t+\Delta t}}{\sqrt{\alpha_{t+\Delta t}}}=\frac{\vx_t}{\sqrt{\alpha_t}}+\left(\sqrt{\frac{1-\alpha_{t+\Delta t}}{\alpha_{t+\Delta t}}}-\sqrt{\frac{1-\alpha_t}{\alpha_t}}\right) \vepsilon_\vtheta\left(\vx_t,t\right).
\end{align*}

\textbf{Latent Diffusion Model~(LDM)}~\cite{rombach2022high} is a variant of text-to-image diffusion models. An autoencoder is first trained on large image datasets, where the encoder $\gE$ compresses the original image $\vx_0$ into a latent code $\vz_{\vx_0}$, and the decoder $\gD$ reconstructs the original image from the latent code. That is 
\begin{align*}
    &\vz_{\vx_0} = \gE(\vx_0), & \vx_0\approx \gD(\vz_{\vx_0}).
\end{align*}
Then a conditional DDPM $p_\vtheta(\vv_{t-1}|\vv_{t},\vc)$ is trained to gradually remove noise for data sampling.
\textbf{Classifier-free guidance~(GFC)}~\cite{ho2021classifier} is proposed to improve the text-image alignment by linearly combining the conditional predicted noise and the unconditional one.
\begin{align*}
\hat{\vepsilon}_\vtheta(\vx_t,t,\vc) = (1+w)\vepsilon_\vtheta(\vx_t,t,\vc)-w\vepsilon_\vtheta(\vx_t,t,\varnothing),
\end{align*}
where $w > 0$ is the guidance scale. Larger $w$ typically improves the image-text alignment but overlarge $w$ causes the degradation of sample fidelity and diversity.
\subsection{Temporal Co-Denoising}\label{sec:temporal co-denoising}

As we discussed above, current Text-to-Video diffusion methods for generation and editing typically view the video as a whole. Given a noisy video $\vv_{t}$, they train a diffusion model $p_\vtheta(\vv_{t-1}|\vv_{t},\vc)$ with respect to noise prediction model $\vepsilon_\vtheta(\vv_{t},t,\vc)$ to denoise it as a whole.
This greatly limits the video length that they are able to generate and makes it hard for them to accommodate multi-text conditions. 
Though some works performed long videos generation via the autoregressive mechanism, this manner suffers from severe content degradation and only supports serialization generation, leading to inference inefficiency. 

In contrast, we consider the denoising process of the entire video as multiple short videos with temporal overlapping undergoing parallel denoising in the temporal domain. We approximate the denoising trajectory of a long video through the joint denoising model of short videos in the temporal domain. More specifically, we suppose there exists a model $p^l_\vtheta(\vv_{t-1}|\vv_{t},\vc)$ (with a corresponding noise prediction network $\epsilon_{\vtheta}^l(\vv_t,t,\vc)$) capable of denoising the given long video $\vv_t$, resulting in the denoising trajectory,
\begin{align*}
 \vv_{T},\vv_{T-1},\dots,\vv_0, \qquad s.t. \quad \vv_{t-1} \sim p^l_\vtheta(\vv_{t-1}|\vv_{t},\vc),
\end{align*}
where we use the diffusion model to gradually transform the pure noise $\vv_T\sim \mathcal N(\boldsymbol{0},\mI)$ to the clean video $\vv_0$. $\vc$ can be represented as a single or multiple text prompts.

We define a set of mappings $F_i$ to project all original videos $\vv_t$ (both noisy and clean) in the trajectory to short video segments $\vv_{t}^i$, specifically,
\begin{align*}
    F_i(\vv_t) = \vv^i_{t} = \vv_{t,S*i:S*i+M}, \quad t=1,2,\dots,T, \quad i=0,1,\dots,N-1,
\end{align*}
where $\vv_{t,S*i-S*i+M}$ represents the collection of frames with frame id from $S*i$ to $S*i+M$, $S$ represents the stride among adjacent short video clips, $M$ is the fixed length of short videos, and $N$ is the total number of clips. Empirically, we find that setting $S$ to $M//2$ or $M//4$ yields excellent results and preserves efficiency. When setting $S=M$, our method degrades into isolated denoising. The total number of frames of the video is $S*N+M$. Each short video $\vv_{t}^i$ is guided with an independent text condition $\vc^i$. After obtaining these short videos, we are able to denoise these short videos using the off-the-shelf short video diffusion models $p_\vtheta^i(\vv^i_{t-1}|\vv^i_{t}, \vc^i)$. For simplicity, we can set the diffusion models for all video clips to a single diffusion model $p_\vtheta^s(\vv_{t-1}^i|\vv_{t}^i,\vc^i)$, and we find that it works quite well. Then we have,
\begin{align*}
    \vv_{t-1}^i\sim p_\vtheta^s(\vv_{t-1}^i| \vv_{t}^i, \vc^i).
\end{align*}
The remaining question is how to obtain the denoised $\vv_{t-1}$ after acquiring all short video clips $\vv_{t-1}^i$.

Considering that we have assumed $F_{i}(\vv_t) = \vv_{t}^i$ for all $t$ and $i$, the ideal $\vv_{t-1}$ should satisfy that $F_{i}(\vv_{t-1})$ is as close as $\vv_{t-1}^i$ as possible. Therefore, the optimal $\vv_{t-1}$ can be obtained by solving the following optimization problem.
\begin{align*}\label{eq:inversion function}
    \vv_{t-1} = \mathop{\arg\min}_{\vv}\sum_{i=0}^{N-1}\left\|W_i \otimes (F_i(\vv)-\vv_{t-1}^i)\ \right\|_2^2,
\end{align*}
where $W_i$ is the pixel-wise weight for the video clip $\vv^i_{t}$, and $\otimes$ means the tensor product. It is not difficult to verify that for an arbitrary frame $j$ in the video $\vv_{t-1}$, namely $\vv_{t-1,j}$, it should be equal to the weighted sum of all the corresponding frames in short videos that contain the $j$ frame. We provide the proof in Sec.~\ref{sec:proof}.

In this way, we are able to approximate the transition function $p_\vtheta^l(\vv_{t-1}|\vv_{t},\vc)$ with $p^s_{\vtheta}(\vv^i_{t-1}|\vv^i_{t},\vc^i)$. As we claimed, our method establishes an abstract long video generator and editor without necessitating any additional training, enabling the generation and editing of videos of any length using established short video generation and editing methodologies.

\textbf{Condition interpolation.}  Considering that it is rather cumbersome to label all short videos with exact text descriptions, we allow the acceptance of sparse conditions. For instance, assuming that only $\vv_{t}^{k i }, k\in \mathbb N_{+}$ are labeled with text descriptions $\vc^{ki}$, we obtain the text conditions of other video clips through adjacent interpolation $\vc^{ki+j} = \frac{k-j}{k}\vc^{ki}+\frac{j}{k}\vc^{k(i+1)}$, where $j=0,1,\dots,k$. This allows for more flexibility in text-based guidance, enabling smoother content generation and simplifying the overall process for users.

\section{Integrate Gen-L-Video with Mainstream Paradigms}\label{sec:application}
As previously mentioned, our method can be applied to three mainstream paradigms: pretrained t2v, tuning-free t2v, and one-shot-tuning t2v. In this section, we will introduce our implementation and improvements for these paradigms in long video generation and editing. Furthermore, by utilizing additional control information, we can achieve more accurate layout control. Advances in open-set detection and segmentation allow us to achieve precise editing of arbitrary objects in the video without altering the other contents~(\eg background), resulting in a more powerful and flexible video editing process. All our implementations are based on the pretrained LDM and its variants. 

\begin{figure*}[t!]
    \centering
    \vspace{-2em}
    \includegraphics[width=\linewidth]{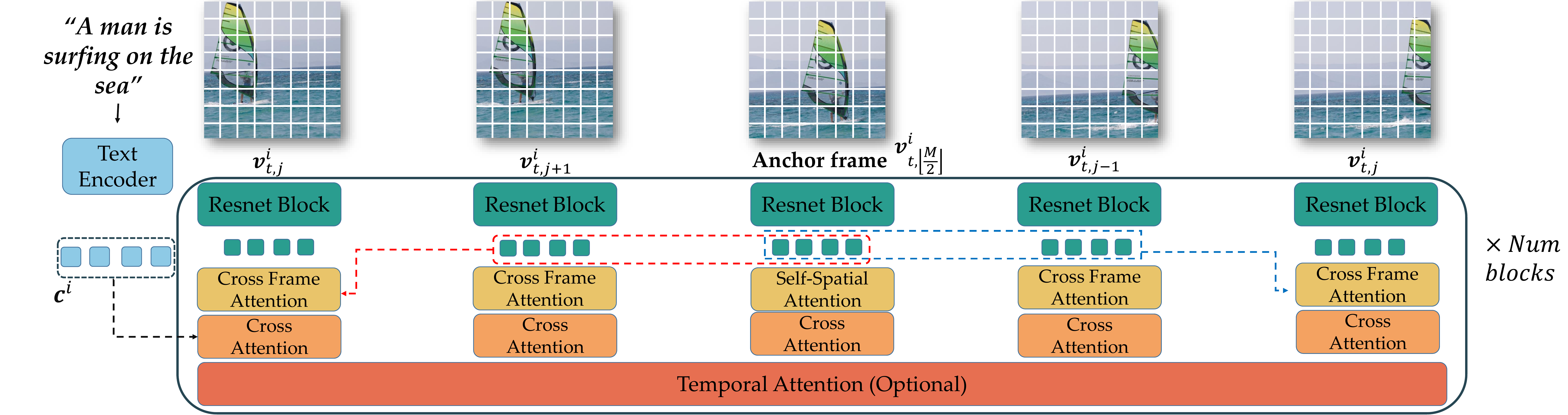}
    \caption{\textit{\textbf{High-level overview of the model architecture for short video denoising.}} The given noisy short video is treated as a collection of noisy images. In each block, we begin by processing each image independently with 2d convolutions (with normalization and residual connection). Then,  spatial attention is applied to the selected anchor frame to enhance pixel interactions, while bi-directional cross-frame attention enables other frames to interact with both adjacent frames and the anchor frame. After that, cross-attention is incorporated to integrate textual information for guiding the denoising process, and temporal attention is trained to further refine the relationships among frames.}
    \label{fig:svdm}
    \vspace{-1em}
\end{figure*}

\paragraph{Pretrained Text-to-Video.}\label{sec:ptv}
For pretrained Text-to-Video generation and editing, we choose the open-sourced VideoCrafter~\cite{he2022latent}. VideoCrafter is a Text-to-Video model fine-tuned from LDM on the large text-video dataset WebVid-10M~\cite{bain2021frozen}. For modeling the dynamic relationships among frames, VideoCrafter adds additional temporal attention blocks in the original LDM. The pipeline for long video generation and editing follows our proposed temporal co-denoising as illustrated in Fig.~\ref{fig:lvdm}. 
\paragraph{Tuning-free Text-to-Video.}\label{sec:tftv}
For tuning-free Text-to-Video, we follow the pipeline of Pix2Video~\cite{ceylan2023pix2video} for its efficiency, with an additional distance penalty among adjacent frames to avoid dramatic changes. Though many tuning-free Text-to-Video generation~\cite{khachatryan2023text2video} or editing~\cite{ceylan2023pix2video,qi2023fatezero} methods apply the sparse causal attention mechanism~\cite{wu2022tune}, where the spatial attention is replaced by the cross attention between each frame and its former adjacent frame and the very first frame in the video. We find that it typically causes noticeable inconsistency between the initial few frames and the subsequent frames when generating long videos as shown in Fig.~\ref{fig:comp_co}. We argue that it is because when the anchor frame is chosen as the first frame, its denoising path will not be influenced by any other frames in the current and following video clips, leading to unidirectional information propagation. It is obvious considering that in sparse causal attention, vanilla spatial attention is conducted in the very first frame, and the denoising of other video clips will also not influence the noise path of the first frame.  Instead, we propose to set the anchor frame as the middle frame in each short video clip and bidirectionally propagate the information to both the start and the end, which we dubbed as \textit{Bi-Directional Cross-Frame Attention}. The bidirectional information propagation allows the mutual influence of anchor frames among different video clips, making it easier to find a compatible denoising path for the long video.  Specifically, denoting the input feature for the cross attention block of $j^{th}$ frame in video $\vv_{t}^i$ as $\vz_{t,j}^i$, the computation of $\mathrm{Attention}(Q,K,V)$ is formulated as 
\begin{align*}
     Q=W^Q \vz_{t,j}^i, \quad K= W^K \vz_{t,j}^{i,*}, \quad V=W^V \vz_{t,j}^{i,*},
\end{align*}
where $\vz_{t,j}^{i,*} = \begin{cases} 
\left[\vz_{t,j-1}^i, \vz_{t,\lfloor M/2 \rfloor}\right], & j>\lfloor M/2 \rfloor\\
\left[\vz_{t,j+1}^i, \vz_{t,\lfloor M/2 \rfloor}^{i}\right], & j < \lfloor M/2 \rfloor\\
\vz_{t,\lfloor M/2 \rfloor}, & j = \lfloor M/2 \rfloor
\end{cases}$ and $[\cdot,\cdot]$ means the operation of sequence dimensional concatenation. 
\paragraph{One-shot tuning Text-to-Video.}\label{sec:osttv}
For one-shot tuning Text-to-Video, we follow the pipeline of Tune-A-Video~\cite{wu2022tune}. Similar to Tuning-free Text-to-Video, we also replace the sparse causal attention mechanism with our proposed bidirectional cross-frame attention. However, applying this pipeline directly to the training and generation of long videos presents non-trivial challenges. Although we can prompt the model to learn denoising for each short video clip, it struggles during generation. This is because many short video clips in a video share the same text description, making it difficult for the model to determine which clip it is denoising based on randomly initialized noise and similar text conditions alone. To address this, we suggest learning clip identifier $\ve^i$ for each short video clip $\vv_{t}^i$ to guide the model in denoising the corresponding clip.

However, introducing $\ve^i$ could lead to overfitting of the corresponding clip content, causing the model to overlook the text information and lose its editing ability. Drawing on the idea of CFG~\cite{ho2021classifier}, we randomly drop $\ve^i$ during training. At test time, we base our approach on:
$$
\hat{\epsilon}_\vtheta(\vv^i_{t},t,\vc^i,\ve^i) = (1+w)  \epsilon_\vtheta(\vv^i_{t},t,\vc^i,\ve^i) - w \epsilon_\vtheta(\vv^i_{t},t,\varnothing,\varnothing).
$$
This effectively alleviates the overfitting phenomenon. We believe that video learning consists of learning \textit{content} and \textit{motion} information. Although $\ve^i$ learns both content and motion information for the video clip $\vv_{t}^i$, we aim to retain only the motion information. By dropping $\ve^i$, the model learns across all video clips, gaining a large amount of content information. Shifting the denoising direction away from the scenario when $\ve^i$ is dropped allows us to reduce overfitting to video content.
\paragraph{Personalized and controllable generation.}
Our method can be easily extended to personalized and controllable layout generation. Users can easily combine personalized diffusion models obtained through fine-tuning strategies like DreamBooth~\cite{ruiz2022dreambooth} and LoRA~\cite{hu2021lora} with our generation pipelines. Besides, we are able to inject additional layout control such as pose and segmentation maps with ControlNet~\cite{zhang2023adding} and T2I-Adapter~\cite{mou2023t2i,ma2023follow} pretrained on image datasets~\cite{laion5b}.  This allows us for more smooth and more precise video generation and editing.
\paragraph{Edit anything in the video.}
Open-set detection~\cite{liu2023grounding} and segmentation~\cite{kirillov2023segment} have demonstrated remarkable ability and inspired plenty of interesting applications. We show that it is possible to combine these with our method to achieve precise arbitrary object editing in long videos. Specifically, given a prompt of a specific object~(\eg man), we apply the open-set detection model Grouding DINO~\cite{liu2023grounding} to detect the corresponding object in each frame. Then, the open-set segmentation model SAM~\cite{kirillov2023segment} is applied to get the precise mask of the object in each frame of the video. With these masks, we are capable of applying inpainting methodologies for video editing while keeping the other components of the video content unchanged. We find that directly using an pretrained Text-to-Image diffusion model with our proposed bi-directional cross-frame attention can already yield acceptable results. To achieve more precise layout generation, we additional add a controlnet pretrained on Text-To-Image datasets~\cite{laion5b} to accept the SAM maps. 
\section{Experiments}
\paragraph{Implementation details.}
All our experiments are heavily built upon the pretrained LDM~\cite{rombach2022high}~(a.k.a Stable Diffusion), as we mentioned. By default, DDIM sampling strategy is applied for all our experiments and the sampling step and guidance scale is set to $50$, and 13.5, respectively. In most cases, we set the number of frames of short video clips to $16$ and stride between adjacent short videos clips to $4$. For the one-shot-tuning Text-to-Video pipeline, we set the basic learning rate as $3e-5$ and scale up the learning rate as the batch size, which greatly accelerates the training. The default beach size is set to 5, and the loss typically converges within 100 epochs for videos with around 100 frames.
\paragraph{Benchmarks.}
To evaluate our method, we collect a video dataset containing 66 videos whose lengths vary from 32 to hundreds of frames. These videos are mostly drawn from the TGVE competition~\cite{wu2022tune} and the internet. For each video, we label it a source prompt and add four edited prompts for video editing, including object change, background change, style transfer, similar motion changes, and multiple changes. We provide details about the dataset in Sec.~\ref{sec:dataset details}.

\begin{figure*}[t!]
    \centering
    \vspace{-4em}
\includegraphics[width=\linewidth]{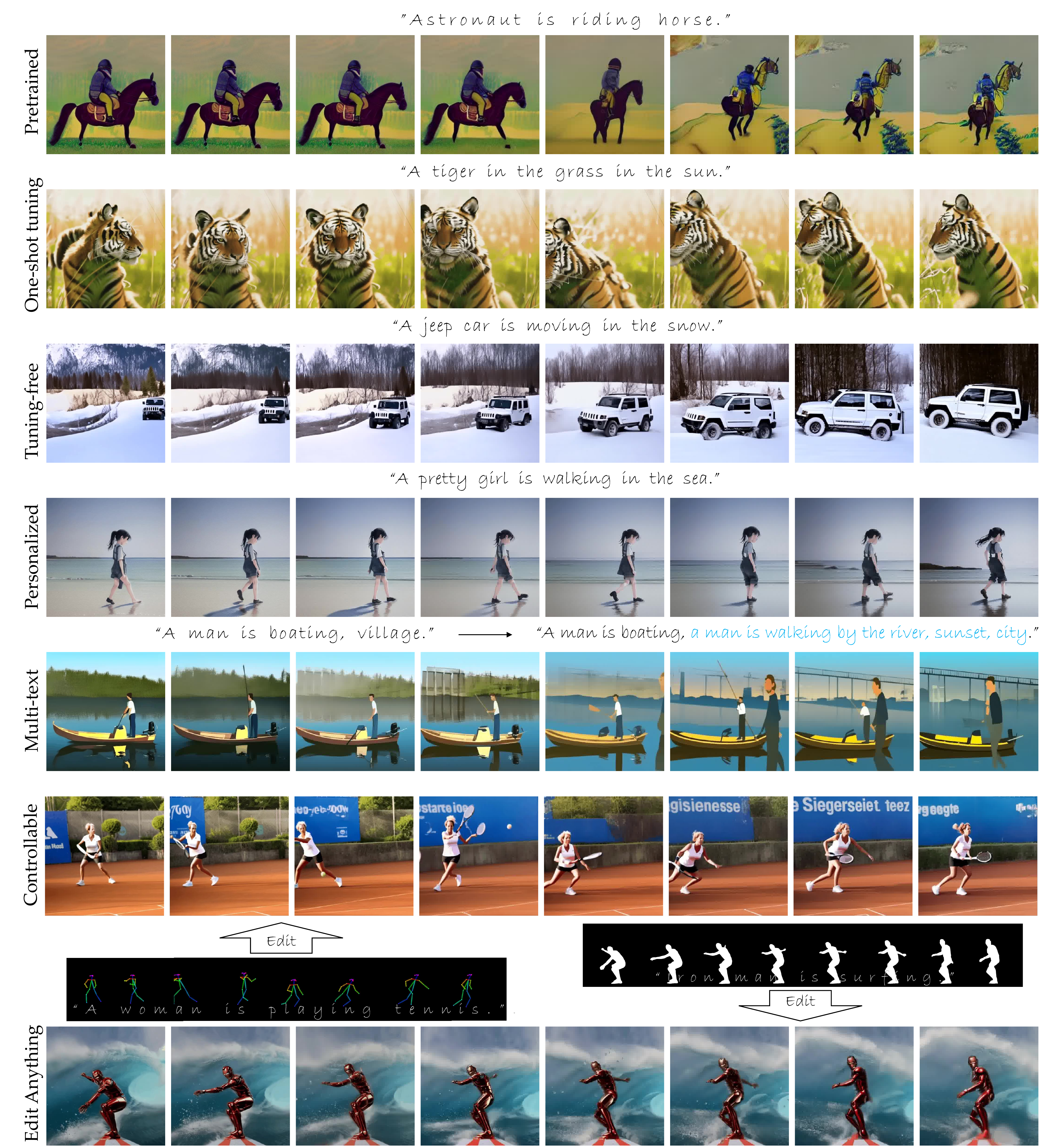}
    \vspace{-1.5em}    \caption{\textit{\textbf{Qualitative generation results of Gen-L-Video.}}}
    \label{fig:qualitative results}
    \vspace{-1.5em}
\end{figure*}

\paragraph{Qualitative results.}
We provide a visual presentation of \textit{representatives} of our generated videos in Fig.~\ref{fig:qualitative results}, including results in various lengths generated through pretrained t2v, tuning-free t2v, one-shot tuning t2v, personalized diffusion model, multi-text conditions, pose layout control, and video inpainting through the auto-detected mask, respectively. All of them show favorable results, demonstrating the strong versatility of our framework. More results can be seen in Sec.~\ref{sec:addtional results} and our project page: \url{https://github.com/G-U-N/Gen-L-Video}.

\begin{table}[t]
\vspace{-2.5em}
\begin{minipage}[t]{0.48\linewidth}
\centering
\caption{\textit{\textbf{Quantitative comparison.}}}
\label{tab:quant}
\vspace{-0.5em}
\resizebox{\textwidth}{!}{%
\begin{tabular}{lccccc}
\toprule
\multirow{2}{*}{Method} & \multicolumn{2}{c}{Frame Consistency}  &  & \multicolumn{2}{c}{Textual Alignment}   \\ \cline{2-3} \cline{5-6} 
          & \textit{Avg. Score} $\uparrow$ & \textit{Human Pref.} $\uparrow$ &  & \textit{Avg. Score} $\uparrow$ & \textit{Variance ($\times$100)} $\downarrow$ \\ \midrule
Isolated  & 91.65      &  14.62\%         &  &  21.16     & 0.57          \\ 
Gen-L-Video  & \textbf{93.18} & \textbf{85.38\%} &  & \textbf{21.18} & \textbf{0.48} \\ \bottomrule
\end{tabular}%
}
\end{minipage}%
\hfill
\begin{minipage}[t]{0.48\linewidth}
\centering
\caption{\textit{\textbf{Comparison to different methods.}}}\label{tab:comparison box}
\vspace{-0.5em}
\resizebox{\textwidth}{!}{%
\begin{tabular}{c c c c c c c}
\toprule
Method & Long & Multi-Text Condition & Vast Video Corpus & Parallel Denoise & Versatile\\ \midrule
Tune-A-Video~\cite{wu2022tune} & \textcolor{red}{\texttimes} & \textcolor{red}{\texttimes} & \textcolor{green}{\texttimes} & \textcolor{red}{\texttimes} & \textcolor{red}{\texttimes} \\
LVDM~\cite{he2022latent} &  \textcolor{green}{\checkmark} & \textcolor{red}{\texttimes} & \textcolor{red}{\checkmark} & \textcolor{red}{\texttimes} & \textcolor{red}{\texttimes}  \\
NUWA-XL~\cite{yin2023nuwa} & \textcolor{green}{\checkmark} & \textcolor{green}{\checkmark} &  \textcolor{red}{\checkmark} &  \textcolor{green}{\checkmark} & \textcolor{red}{\texttimes} \\ \midrule
Gen-L-Video & \textcolor{green}{\checkmark} & \textcolor{green}{\checkmark} & \textcolor{green}{\texttimes} & \textcolor{green}{\checkmark} &  \textcolor{green}{\checkmark} \\
\bottomrule
\end{tabular}
}

\label{discussion_table}
\end{minipage}
\vspace{-0.5em}
\end{table}

\begin{figure*}[t!]
    \centering
    \includegraphics[width=\linewidth]{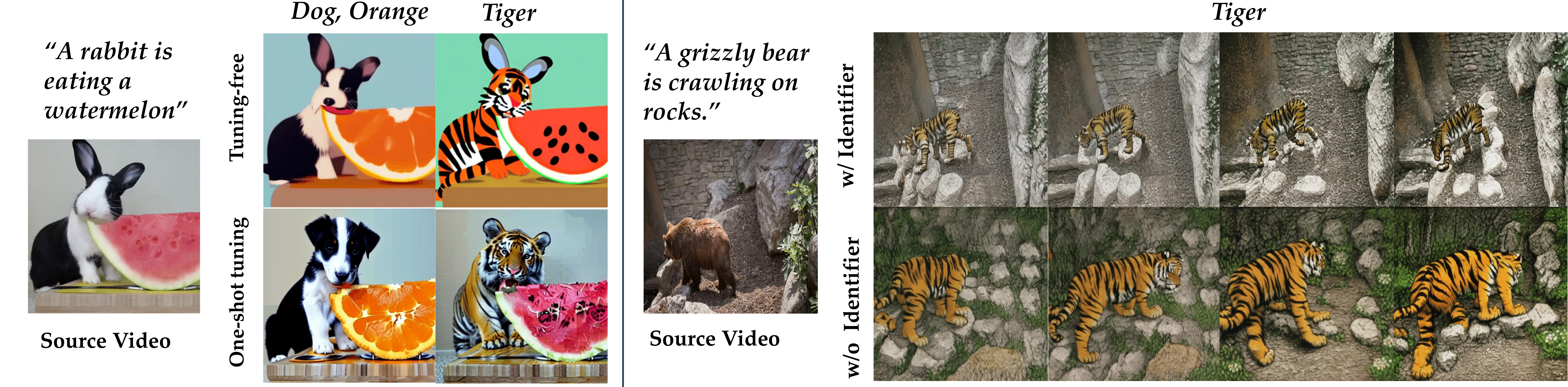}
    \caption{\textit{\textbf{Left: Comparison between tuning-free t2v and one-shot tuning t2v.}} One-shot tuning t2v yields more editing flexibility. \textit{\textbf{Right: Comparison between one-shot-tuning t2v w/ or w/o clip identifier.}} One-shot-tuning t2v  w/o clip identifier fails to generate consecutive motions in the source video with random initial noise.}
    \label{fig:comp}
    \vspace{-1em}
\end{figure*}

\paragraph{Quantitative results.}
For quantitative comparisons, we assess the video frame consistency and the textual alignment following the prior work~\cite{wu2022tune}. We compare Gen-L-Video with \textit{Isolated Denoising}, where each video clip is denoised isolatedly. Regarding the \textit{video frame consistency}, we employ the CLIP~\cite{radford2021learning} image encoder to extract the embeddings of individual frames and then compute the average cosine similarity between all pairs of video frames. To evaluate the \textit{textual alignment}, which refers to the alignment between text and video, we calculate the CLIP score of text and each frame in the video. The average value of the scores is used to measure the alignment degree while the variance of those is used to measure the alignment stability. For human preference, we select several participants to vote on which method yields better frame consistency and textual alignment and get 1040 votes in total. The experimental results are shown in Table.~\ref{tab:quant}.

\vspace{-3mm}
\paragraph{Ablation study.}
\textit{Bi-directional cross-frame attention.} We compare our proposed Bi-directional cross-frame attention with the sparse causal attention when temporal co-denoising is applied. As illustrated in Fig.~\ref{fig:comp_co},  our method achieves the most smooth and consistent result while sparse causal attention typically causes the first few frames incwonsistent with subsequent ones. \textit{Video clip identifier}. We compare the generation results of one-shot tuning t2v with or without the clip identifier in Fig.~\ref{fig:comp}~(Right). When random initial noise is used, one-shot tuning t2v without the clip identifier fails to generate consecutive content in the source video, while the other method succeeds

\vspace{-3mm}
\section{Conclusion}
\vspace{-3mm}
In this work, we propose \textit{Gen-L-Video}, a universal methodology that extends short video diffusion models for efficient multi-text conditioned long video generation and editing. We implement mainstream Text-to-Video methods and make additional improvements to integrate them with  \textit{Gen-L-Video} for long video generation and editing. Experiments verify our framework is universal and scalable.

\noindent\textbf{Limitations}: In general, our framework should be able to be extended into the co-working of various different video diffusion models with different lengths to obtain more flexibility in generation and editing, but we haven't experimented with that. This avenue of research is left as future work.

\bibliographystyle{plainnat}
\bibliography{refs}

\begin{thebibliography}{74}
\providecommand{\natexlab}[1]{#1}
\providecommand{\url}[1]{\texttt{#1}}
\expandafter\ifx\csname urlstyle\endcsname\relax
  \providecommand{\doi}[1]{doi: #1}\else
  \providecommand{\doi}{doi: \begingroup \urlstyle{rm}\Url}\fi

\bibitem[Arora and Lee(2021)]{arora2021singan-gif}
Rajat Arora and Yong~Jae Lee.
\newblock Singan-gif: Learning a generative video model from a single gif.
\newblock In \emph{CVPR}, pages 1310--1319, 2021.

\bibitem[Bain et~al.(2021)Bain, Nagrani, Varol, and Zisserman]{bain2021frozen}
Max Bain, Arsha Nagrani, G{\"u}l Varol, and Andrew Zisserman.
\newblock Frozen in time: A joint video and image encoder for end-to-end
  retrieval.
\newblock In \emph{ICCV}, pages 1728--1738, 2021.

\bibitem[Ceylan et~al.(2023)Ceylan, Huang, and Mitra]{ceylan2023pix2video}
Duygu Ceylan, Chun-Hao~Paul Huang, and Niloy~J Mitra.
\newblock Pix2video: Video editing using image diffusion.
\newblock \emph{arXiv preprint arXiv:2303.12688}, 2023.

\bibitem[Couairon et~al.(2022)Couairon, Verbeek, Schwenk, and
  Cord]{couairon2022diffedit}
Guillaume Couairon, Jakob Verbeek, Holger Schwenk, and Matthieu Cord.
\newblock Diffedit: Diffusion-based semantic image editing with mask guidance.
\newblock \emph{arXiv preprint arXiv:2210.11427}, 2022.

\bibitem[Ding et~al.(2022)Ding, Zheng, Hong, and Tang]{ding2022cogview2}
Ming Ding, Wendi Zheng, Wenyi Hong, and Jie Tang.
\newblock Cogview2: Faster and better text-to-image generation via hierarchical
  transformers.
\newblock \emph{arXiv preprint arXiv:2204.14217}, 2022.

\bibitem[Esser et~al.(2023)Esser, Chiu, Atighehchian, Granskog, and
  Germanidis]{esser2023structure}
Patrick Esser, Johnathan Chiu, Parmida Atighehchian, Jonathan Granskog, and
  Anastasis Germanidis.
\newblock Structure and content-guided video synthesis with diffusion models.
\newblock \emph{arXiv preprint arXiv:2302.03011}, 2023.

\bibitem[Gafni et~al.(2022)Gafni, Polyak, Ashual, Sheynin, Parikh, and
  Taigman]{gafni2022make}
Oran Gafni, Adam Polyak, Oron Ashual, Shelly Sheynin, Devi Parikh, and Yaniv
  Taigman.
\newblock Make-a-scene: Scene-based text-to-image generation with human priors.
\newblock In \emph{ECCV}, pages 89--106. Springer, 2022.

\bibitem[Gal et~al.(2022)Gal, Alaluf, Atzmon, Patashnik, Bermano, Chechik, and
  Cohen-Or]{gal2022image}
Rinon Gal, Yuval Alaluf, Yuval Atzmon, Or~Patashnik, Amit~H Bermano, Gal
  Chechik, and Daniel Cohen-Or.
\newblock An image is worth one word: Personalizing text-to-image generation
  using textual inversion.
\newblock \emph{arXiv preprint arXiv:2208.01618}, 2022.

\bibitem[Ge et~al.(2022)Ge, Hayes, Yang, Yin, Pang, Jacobs, Huang, and
  Parikh]{ge2022long}
Songwei Ge, Thomas Hayes, Harry Yang, Xi~Yin, Guan Pang, David Jacobs, Jia-Bin
  Huang, and Devi Parikh.
\newblock Long video generation with time-agnostic vqgan and time-sensitive
  transformer.
\newblock \emph{arXiv preprint arXiv:2204.03638}, 2022.

\bibitem[Goodfellow et~al.(2020)Goodfellow, Pouget-Abadie, Mirza, Xu,
  Warde-Farley, Ozair, Courville, and Bengio]{goodfellow2020generative}
Ian Goodfellow, Jean Pouget-Abadie, Mehdi Mirza, Bing Xu, David Warde-Farley,
  Sherjil Ozair, Aaron Courville, and Yoshua Bengio.
\newblock Generative adversarial networks.
\newblock \emph{Communications of the ACM}, 63\penalty0 (11):\penalty0
  139--144, 2020.

\bibitem[Gu et~al.(2022)Gu, Chen, Bao, Wen, Zhang, Chen, Yuan, and
  Guo]{gu2022vector}
Shuyang Gu, Dong Chen, Jianmin Bao, Fang Wen, Bo~Zhang, Dongdong Chen, Lu~Yuan,
  and Baining Guo.
\newblock Vector quantized diffusion model for text-to-image synthesis.
\newblock In \emph{CVPR}, pages 10696--10706, 2022.

\bibitem[Gupta et~al.(2018)Gupta, Schwenk, Farhadi, Hoiem, and
  Kembhavi]{gupta2018imagine}
Tanmay Gupta, Dustin Schwenk, Ali Farhadi, Derek Hoiem, and Aniruddha Kembhavi.
\newblock Imagine this! scripts to compositions to videos.
\newblock In \emph{ECCV}, pages 598--613, 2018.

\bibitem[Gur et~al.(2020)Gur, Benaim, and Wolf]{gur2020hierarchical}
Shir Gur, Sagie Benaim, and Lior Wolf.
\newblock Hierarchical patch vae-gan: Generating diverse videos from a single
  sample.
\newblock \emph{NeurIPS}, 33:\penalty0 16761--16772, 2020.

\bibitem[Harvey et~al.(2022)Harvey, Naderiparizi, Masrani, Weilbach, and
  Wood]{harvey2022flexible}
William Harvey, Saeid Naderiparizi, Vaden Masrani, Christian Weilbach, and
  Frank Wood.
\newblock Flexible diffusion modeling of long videos.
\newblock \emph{arXiv preprint arXiv:2205.11495}, 2022.

\bibitem[He et~al.(2022)He, Yang, Zhang, Shan, and Chen]{he2022latent}
Yingqing He, Tianyu Yang, Yong Zhang, Ying Shan, and Qifeng Chen.
\newblock Latent video diffusion models for high-fidelity video generation with
  arbitrary lengths.
\newblock \emph{arXiv preprint arXiv:2211.13221}, 2022.

\bibitem[Hertz et~al.(2022)Hertz, Mokady, Tenenbaum, Aberman, Pritch, and
  Cohen-Or]{hertz2022prompt}
Amir Hertz, Ron Mokady, Jay Tenenbaum, Kfir Aberman, Yael Pritch, and Daniel
  Cohen-Or.
\newblock Prompt-to-prompt image editing with cross attention control.
\newblock \emph{arXiv preprint arXiv:2208.01626}, 2022.

\bibitem[Ho and Salimans(2021)]{ho2021classifier}
Jonathan Ho and Tim Salimans.
\newblock Classifier-free diffusion guidance.
\newblock In \emph{NeurIPS}, 2021.

\bibitem[Ho and Salimans(2022)]{ho2022classifier}
Jonathan Ho and Tim Salimans.
\newblock Classifier-free diffusion guidance.
\newblock \emph{arXiv preprint arXiv:2207.12598}, 2022.

\bibitem[Ho et~al.(2020)Ho, Jain, and Abbeel]{ho2020denoising}
Jonathan Ho, Ajay Jain, and Pieter Abbeel.
\newblock Denoising diffusion probabilistic models.
\newblock \emph{Advances in Neural Information Processing Systems},
  33:\penalty0 6840--6851, 2020.

\bibitem[Ho et~al.(2022{\natexlab{a}})Ho, Chan, Saharia, Whang, Gao, Gritsenko,
  Kingma, Poole, Norouzi, Fleet, et~al.]{ho2022imagen}
Jonathan Ho, William Chan, Chitwan Saharia, Jay Whang, Ruiqi Gao, Alexey
  Gritsenko, Diederik~P Kingma, Ben Poole, Mohammad Norouzi, David~J Fleet,
  et~al.
\newblock Imagen video: High definition video generation with diffusion models.
\newblock \emph{arXiv preprint arXiv:2210.02303}, 2022{\natexlab{a}}.

\bibitem[Ho et~al.(2022{\natexlab{b}})Ho, Salimans, Gritsenko, Chan, Norouzi,
  and Fleet]{ho2022video}
Jonathan Ho, Tim Salimans, Alexey Gritsenko, William Chan, Mohammad Norouzi,
  and David~J Fleet.
\newblock Video diffusion models.
\newblock \emph{arXiv:2204.03458}, 2022{\natexlab{b}}.

\bibitem[Hong et~al.(2022)Hong, Ding, Zheng, Liu, and Tang]{hong2022cogvideo}
Wenyi Hong, Ming Ding, Wendi Zheng, Xinghan Liu, and Jie Tang.
\newblock Cogvideo: Large-scale pretraining for text-to-video generation via
  transformers.
\newblock \emph{arXiv preprint arXiv:2205.15868}, 2022.

\bibitem[Hu et~al.(2021)Hu, Shen, Wallis, Allen-Zhu, Li, Wang, Wang, and
  Chen]{hu2021lora}
Edward~J Hu, Yelong Shen, Phillip Wallis, Zeyuan Allen-Zhu, Yuanzhi Li, Shean
  Wang, Lu~Wang, and Weizhu Chen.
\newblock Lora: Low-rank adaptation of large language models.
\newblock \emph{arXiv preprint arXiv:2106.09685}, 2021.

\bibitem[Kawar et~al.(2022)Kawar, Zada, Lang, Tov, Chang, Dekel, Mosseri, and
  Irani]{kawar2022imagic}
Bahjat Kawar, Shiran Zada, Oran Lang, Omer Tov, Huiwen Chang, Tali Dekel, Inbar
  Mosseri, and Michal Irani.
\newblock Imagic: Text-based real image editing with diffusion models.
\newblock \emph{arXiv preprint arXiv:2210.09276}, 2022.

\bibitem[Khachatryan et~al.(2023)Khachatryan, Movsisyan, Tadevosyan, Henschel,
  Wang, Navasardyan, and Shi]{khachatryan2023text2video}
Levon Khachatryan, Andranik Movsisyan, Vahram Tadevosyan, Roberto Henschel,
  Zhangyang Wang, Shant Navasardyan, and Humphrey Shi.
\newblock Text2video-zero: Text-to-image diffusion models are zero-shot video
  generators.
\newblock \emph{arXiv preprint arXiv:2303.13439}, 2023.

\bibitem[Kirillov et~al.(2023)Kirillov, Mintun, Ravi, Mao, Rolland, Gustafson,
  Xiao, Whitehead, Berg, Lo, et~al.]{kirillov2023segment}
Alexander Kirillov, Eric Mintun, Nikhila Ravi, Hanzi Mao, Chloe Rolland, Laura
  Gustafson, Tete Xiao, Spencer Whitehead, Alexander~C Berg, Wan-Yen Lo, et~al.
\newblock Segment anything.
\newblock \emph{arXiv preprint arXiv:2304.02643}, 2023.

\bibitem[Li et~al.(2018)Li, Min, Shen, Carlson, and Carin]{li2018video}
Yitong Li, Martin Min, Dinghan Shen, David Carlson, and Lawrence Carin.
\newblock Video generation from text.
\newblock In \emph{AAAI}, volume~32, 2018.

\bibitem[Liang et~al.(2022)Liang, Wu, Hu, Gan, Wang, Wang, Liu, Fang, and
  Duan]{liang2022nuwa}
Jian Liang, Chenfei Wu, Xiaowei Hu, Zhe Gan, Jianfeng Wang, Lijuan Wang,
  Zicheng Liu, Yuejian Fang, and Nan Duan.
\newblock Nuwa-infinity: Autoregressive over autoregressive generation for
  infinite visual synthesis.
\newblock \emph{NeurIPS}, 35:\penalty0 15420--15432, 2022.

\bibitem[Liu et~al.(2023{\natexlab{a}})Liu, Zhang, Li, Lin, and
  Jia]{liu2023video}
Shaoteng Liu, Yuechen Zhang, Wenbo Li, Zhe Lin, and Jiaya Jia.
\newblock Video-p2p: Video editing with cross-attention control.
\newblock \emph{arXiv preprint arXiv:2303.04761}, 2023{\natexlab{a}}.

\bibitem[Liu et~al.(2023{\natexlab{b}})Liu, Zeng, Ren, Li, Zhang, Yang, Li,
  Yang, Su, Zhu, et~al.]{liu2023grounding}
Shilong Liu, Zhaoyang Zeng, Tianhe Ren, Feng Li, Hao Zhang, Jie Yang, Chunyuan
  Li, Jianwei Yang, Hang Su, Jun Zhu, et~al.
\newblock Grounding dino: Marrying dino with grounded pre-training for open-set
  object detection.
\newblock \emph{arXiv preprint arXiv:2303.05499}, 2023{\natexlab{b}}.

\bibitem[Liu et~al.(2019)Liu, Wang, Yuan, and Zhu]{liu2019cross}
Yue Liu, Xin Wang, Yitian Yuan, and Wenwu Zhu.
\newblock Cross-modal dual learning for sentence-to-video generation.
\newblock In \emph{ACM MM}, pages 1239--1247, 2019.

\bibitem[Luo et~al.(2023)Luo, Chen, Zhang, Huang, Wang, Shen, Zhao, Zhou, and
  Tan]{luo2023videofusion}
Zhengxiong Luo, Dayou Chen, Yingya Zhang, Yan Huang, Liang Wang, Yujun Shen,
  Deli Zhao, Jingren Zhou, and Tieniu Tan.
\newblock Videofusion: Decomposed diffusion models for high-quality video
  generation.
\newblock \emph{arXiv e-prints}, pages arXiv--2303, 2023.

\bibitem[Ma et~al.(2023)Ma, He, Cun, Wang, Shan, Li, and Chen]{ma2023follow}
Yue Ma, Yingqing He, Xiaodong Cun, Xintao Wang, Ying Shan, Xiu Li, and Qifeng
  Chen.
\newblock Follow your pose: Pose-guided text-to-video generation using
  pose-free videos.
\newblock \emph{arXiv preprint arXiv:2304.01186}, 2023.

\bibitem[Marwah et~al.(2017)Marwah, Mittal, and
  Balasubramanian]{marwah2017attentive}
Tanya Marwah, Gaurav Mittal, and Vineeth~N Balasubramanian.
\newblock Attentive semantic video generation using captions.
\newblock In \emph{ICCV}, pages 1426--1434, 2017.

\bibitem[Meng et~al.(2021)Meng, He, Song, Song, Wu, Zhu, and
  Ermon]{meng2021sdedit}
Chenlin Meng, Yutong He, Yang Song, Jiaming Song, Jiajun Wu, Jun-Yan Zhu, and
  Stefano Ermon.
\newblock Sdedit: Guided image synthesis and editing with stochastic
  differential equations.
\newblock In \emph{ICLR}, 2021.

\bibitem[Mittal et~al.(2017)Mittal, Marwah, and
  Balasubramanian]{mittal2017sync}
Gaurav Mittal, Tanya Marwah, and Vineeth~N Balasubramanian.
\newblock Sync-draw: Automatic video generation using deep recurrent attentive
  architectures.
\newblock In \emph{ACM MM}, pages 1096--1104, 2017.

\bibitem[Mokady et~al.(2022)Mokady, Hertz, Aberman, Pritch, and
  Cohen-Or]{mokady2022null}
Ron Mokady, Amir Hertz, Kfir Aberman, Yael Pritch, and Daniel Cohen-Or.
\newblock Null-text inversion for editing real images using guided diffusion
  models.
\newblock \emph{arXiv preprint arXiv:2211.09794}, 2022.

\bibitem[Molad et~al.(2023)Molad, Horwitz, Valevski, Acha, Matias, Pritch,
  Leviathan, and Hoshen]{molad2023dreamix}
Eyal Molad, Eliahu Horwitz, Dani Valevski, Alex~Rav Acha, Yossi Matias, Yael
  Pritch, Yaniv Leviathan, and Yedid Hoshen.
\newblock Dreamix: Video diffusion models are general video editors.
\newblock \emph{arXiv preprint arXiv:2302.01329}, 2023.

\bibitem[Mou et~al.(2023)Mou, Wang, Xie, Zhang, Qi, Shan, and Qie]{mou2023t2i}
Chong Mou, Xintao Wang, Liangbin Xie, Jian Zhang, Zhongang Qi, Ying Shan, and
  Xiaohu Qie.
\newblock T2i-adapter: Learning adapters to dig out more controllable ability
  for text-to-image diffusion models.
\newblock \emph{arXiv preprint arXiv:2302.08453}, 2023.

\bibitem[Nichol et~al.(2021)Nichol, Dhariwal, Ramesh, Shyam, Mishkin, McGrew,
  Sutskever, and Chen]{nichol2021glide}
Alex Nichol, Prafulla Dhariwal, Aditya Ramesh, Pranav Shyam, Pamela Mishkin,
  Bob McGrew, Ilya Sutskever, and Mark Chen.
\newblock Glide: Towards photorealistic image generation and editing with
  text-guided diffusion models.
\newblock \emph{arXiv preprint arXiv:2112.10741}, 2021.

\bibitem[Nikankin et~al.(2022)Nikankin, Haim, and Irani]{nikankin2022sinfusion}
Yaniv Nikankin, Niv Haim, and Michal Irani.
\newblock Sinfusion: Training diffusion models on a single image or video.
\newblock \emph{arXiv preprint arXiv:2211.11743}, 2022.

\bibitem[Pan et~al.(2017)Pan, Qiu, Yao, Li, and Mei]{pan2017create}
Yingwei Pan, Zhaofan Qiu, Ting Yao, Houqiang Li, and Tao Mei.
\newblock To create what you tell: Generating videos from captions.
\newblock In \emph{ACM MM}, pages 1789--1798, 2017.

\bibitem[Qi et~al.(2023)Qi, Cun, Zhang, Lei, Wang, Shan, and
  Chen]{qi2023fatezero}
Chenyang Qi, Xiaodong Cun, Yong Zhang, Chenyang Lei, Xintao Wang, Ying Shan,
  and Qifeng Chen.
\newblock Fatezero: Fusing attentions for zero-shot text-based video editing.
\newblock \emph{arXiv preprint arXiv:2303.09535}, 2023.

\bibitem[Radford et~al.(2021)Radford, Kim, Hallacy, Ramesh, Goh, Agarwal,
  Sastry, Askell, Mishkin, Clark, et~al.]{radford2021learning}
Alec Radford, Jong~Wook Kim, Chris Hallacy, Aditya Ramesh, Gabriel Goh,
  Sandhini Agarwal, Girish Sastry, Amanda Askell, Pamela Mishkin, Jack Clark,
  et~al.
\newblock Learning transferable visual models from natural language
  supervision.
\newblock In \emph{ICML}, pages 8748--8763. PMLR, 2021.

\bibitem[Ramesh et~al.(2021)Ramesh, Pavlov, Goh, Gray, Voss, Radford, Chen, and
  Sutskever]{ramesh2021zero}
Aditya Ramesh, Mikhail Pavlov, Gabriel Goh, Scott Gray, Chelsea Voss, Alec
  Radford, Mark Chen, and Ilya Sutskever.
\newblock Zero-shot text-to-image generation.
\newblock In \emph{ICML}, pages 8821--8831. PMLR, 2021.

\bibitem[Ramesh et~al.(2022)Ramesh, Dhariwal, Nichol, Chu, and
  Chen]{ramesh2022hierarchical}
Aditya Ramesh, Prafulla Dhariwal, Alex Nichol, Casey Chu, and Mark Chen.
\newblock Hierarchical text-conditional image generation with clip latents.
\newblock \emph{arXiv preprint arXiv:2204.06125}, 2022.

\bibitem[Rombach et~al.(2022)Rombach, Blattmann, Lorenz, Esser, and
  Ommer]{rombach2022high}
Robin Rombach, Andreas Blattmann, Dominik Lorenz, Patrick Esser, and Bj{\"o}rn
  Ommer.
\newblock High-resolution image synthesis with latent diffusion models.
\newblock In \emph{CVPR}, pages 10684--10695, 2022.

\bibitem[Ruiz et~al.(2022)Ruiz, Li, Jampani, Pritch, Rubinstein, and
  Aberman]{ruiz2022dreambooth}
Nataniel Ruiz, Yuanzhen Li, Varun Jampani, Yael Pritch, Michael Rubinstein, and
  Kfir Aberman.
\newblock Dreambooth: Fine tuning text-to-image diffusion models for
  subject-driven generation.
\newblock \emph{arXiv preprint arXiv:2208.12242}, 2022.

\bibitem[Saharia et~al.(2022)Saharia, Chan, Saxena, Li, Whang, Denton,
  Ghasemipour, Ayan, Mahdavi, Lopes, et~al.]{saharia2022photorealistic}
Chitwan Saharia, William Chan, Saurabh Saxena, Lala Li, Jay Whang, Emily
  Denton, Seyed Kamyar~Seyed Ghasemipour, Burcu~Karagol Ayan, S~Sara Mahdavi,
  Rapha~Gontijo Lopes, et~al.
\newblock Photorealistic text-to-image diffusion models with deep language
  understanding.
\newblock \emph{arXiv preprint arXiv:2205.11487}, 2022.

\bibitem[Schuhmann et~al.(2022{\natexlab{a}})Schuhmann, Beaumont, Vencu,
  Gordon, Wightman, Cherti, Coombes, Katta, Mullis, Wortsman,
  et~al.]{schuhmann2022laion}
Christoph Schuhmann, Romain Beaumont, Richard Vencu, Cade Gordon, Ross
  Wightman, Mehdi Cherti, Theo Coombes, Aarush Katta, Clayton Mullis, Mitchell
  Wortsman, et~al.
\newblock Laion-5b: An open large-scale dataset for training next generation
  image-text models.
\newblock \emph{arXiv preprint arXiv:2210.08402}, 2022{\natexlab{a}}.

\bibitem[Schuhmann et~al.(2022{\natexlab{b}})Schuhmann, Vencu, Beaumont,
  Coombes, Gordon, Katta, Kaczmarczyk, and Jitsev]{laion5b}
Christoph Schuhmann, Richard Vencu, Romain Beaumont, Theo Coombes, Cade Gordon,
  Aarush Katta, Robert Kaczmarczyk, and Jenia Jitsev.
\newblock {LAION-5B:} laion-5b: A new era of open large-scale multi-modal
  datasets.
\newblock
  \url{https://laion.ai/laion-5b-a-new-era-of-open-large-scale-multi-modal-datasets/},
  2022{\natexlab{b}}.

\bibitem[Singer et~al.(2022)Singer, Polyak, Hayes, Yin, An, Zhang, Hu, Yang,
  Ashual, Gafni, et~al.]{singer2022make}
Uriel Singer, Adam Polyak, Thomas Hayes, Xi~Yin, Jie An, Songyang Zhang, Qiyuan
  Hu, Harry Yang, Oron Ashual, Oran Gafni, et~al.
\newblock Make-a-video: Text-to-video generation without text-video data.
\newblock \emph{arXiv preprint arXiv:2209.14792}, 2022.

\bibitem[Smith et~al.(2023)Smith, Hsu, Zhang, Hua, Kira, Shen, and
  Jin]{smith2023continual}
James~Seale Smith, Yen-Chang Hsu, Lingyu Zhang, Ting Hua, Zsolt Kira, Yilin
  Shen, and Hongxia Jin.
\newblock Continual diffusion: Continual customization of text-to-image
  diffusion with c-lora.
\newblock \emph{arXiv preprint arXiv:2304.06027}, 2023.

\bibitem[Song et~al.(2020)Song, Meng, and Ermon]{song2020denoising}
Jiaming Song, Chenlin Meng, and Stefano Ermon.
\newblock Denoising diffusion implicit models.
\newblock \emph{arXiv preprint arXiv:2010.02502}, 2020.

\bibitem[Tao et~al.(2020)Tao, Tang, Wu, Sebe, Jing, Wu, and Bao]{tao2020df}
Ming Tao, Hao Tang, Songsong Wu, Nicu Sebe, Xiao-Yuan Jing, Fei Wu, and Bingkun
  Bao.
\newblock Df-gan: Deep fusion generative adversarial networks for text-to-image
  synthesis.
\newblock \emph{arXiv preprint arXiv:2008.05865}, 2020.

\bibitem[Tumanyan et~al.(2022)Tumanyan, Geyer, Bagon, and
  Dekel]{tumanyan2022plug}
Narek Tumanyan, Michal Geyer, Shai Bagon, and Tali Dekel.
\newblock Plug-and-play diffusion features for text-driven image-to-image
  translation.
\newblock \emph{arXiv preprint arXiv:2211.12572}, 2022.

\bibitem[Van Den~Oord et~al.(2017)Van Den~Oord, Vinyals, et~al.]{van2017neural}
Aaron Van Den~Oord, Oriol Vinyals, et~al.
\newblock Neural discrete representation learning.
\newblock \emph{NeurIPS}, 30, 2017.

\bibitem[Villegas et~al.(2022)Villegas, Babaeizadeh, Kindermans, Moraldo,
  Zhang, Saffar, Castro, Kunze, and Erhan]{villegas2022phenaki}
Ruben Villegas, Mohammad Babaeizadeh, Pieter-Jan Kindermans, Hernan Moraldo,
  Han Zhang, Mohammad~Taghi Saffar, Santiago Castro, Julius Kunze, and Dumitru
  Erhan.
\newblock Phenaki: Variable length video generation from open domain textual
  description.
\newblock \emph{arXiv preprint arXiv:2210.02399}, 2022.

\bibitem[Voleti et~al.(2022)Voleti, Jolicoeur-Martineau, and
  Pal]{voleti2022masked}
Vikram Voleti, Alexia Jolicoeur-Martineau, and Christopher Pal.
\newblock Masked conditional video diffusion for prediction, generation, and
  interpolation.
\newblock \emph{arXiv preprint arXiv:2205.09853}, 2022.

\bibitem[Wang et~al.(2022)Wang, Zhou, Ye, and Zhan]{wang2022foster}
Fu-Yun Wang, Da-Wei Zhou, Han-Jia Ye, and De-Chuan Zhan.
\newblock Foster: Feature boosting and compression for class-incremental
  learning.
\newblock In \emph{ECCV}, pages 398--414. Springer, 2022.

\bibitem[Wu and De~la Torre(2022)]{wu2022unifying}
Chen~Henry Wu and Fernando De~la Torre.
\newblock Unifying diffusion models' latent space, with applications to
  cyclediffusion and guidance.
\newblock \emph{arXiv preprint arXiv:2210.05559}, 2022.

\bibitem[Wu et~al.(2022{\natexlab{a}})Wu, Liang, Ji, Yang, Fang, Jiang, and
  Duan]{wu2022nuwa}
Chenfei Wu, Jian Liang, Lei Ji, Fan Yang, Yuejian Fang, Daxin Jiang, and Nan
  Duan.
\newblock N{\"u}wa: Visual synthesis pre-training for neural visual world
  creation.
\newblock In \emph{ECCV}, pages 720--736. Springer, 2022{\natexlab{a}}.

\bibitem[Wu et~al.(2022{\natexlab{b}})Wu, Ge, Wang, Lei, Gu, Hsu, Shan, Qie,
  and Shou]{wu2022tune}
Jay~Zhangjie Wu, Yixiao Ge, Xintao Wang, Weixian Lei, Yuchao Gu, Wynne Hsu,
  Ying Shan, Xiaohu Qie, and Mike~Zheng Shou.
\newblock Tune-a-video: One-shot tuning of image diffusion models for
  text-to-video generation.
\newblock \emph{arXiv preprint arXiv:2212.11565}, 2022{\natexlab{b}}.

\bibitem[Xu et~al.(2018)Xu, Zhang, Huang, Zhang, Gan, Huang, and
  He]{xu2018attngan}
Tao Xu, Pengchuan Zhang, Qiuyuan Huang, Han Zhang, Zhe Gan, Xiaolei Huang, and
  Xiaodong He.
\newblock Attngan: Fine-grained text to image generation with attentional
  generative adversarial networks.
\newblock In \emph{CVPR}, pages 1316--1324, 2018.

\bibitem[Ye et~al.(2021)Ye, Yang, Takac, Sunderraman, and Ji]{ye2021improving}
Hui Ye, Xiulong Yang, Martin Takac, Rajshekhar Sunderraman, and Shihao Ji.
\newblock Improving text-to-image synthesis using contrastive learning.
\newblock \emph{arXiv preprint arXiv:2107.02423}, 2021.

\bibitem[Yin et~al.(2023)Yin, Wu, Yang, Wang, Wang, Ni, Yang, Li, Liu, Yang,
  et~al.]{yin2023nuwa}
Shengming Yin, Chenfei Wu, Huan Yang, Jianfeng Wang, Xiaodong Wang, Minheng Ni,
  Zhengyuan Yang, Linjie Li, Shuguang Liu, Fan Yang, et~al.
\newblock Nuwa-xl: Diffusion over diffusion for extremely long video
  generation.
\newblock \emph{arXiv preprint arXiv:2303.12346}, 2023.

\bibitem[Yu et~al.(2021)Yu, Li, Koh, Zhang, Pang, Qin, Ku, Xu, Baldridge, and
  Wu]{yu2021vector}
Jiahui Yu, Xin Li, Jing~Yu Koh, Han Zhang, Ruoming Pang, James Qin, Alexander
  Ku, Yuanzhong Xu, Jason Baldridge, and Yonghui Wu.
\newblock Vector-quantized image modeling with improved vqgan.
\newblock \emph{arXiv preprint arXiv:2110.04627}, 2021.

\bibitem[Yu et~al.(2022)Yu, Xu, Koh, Luong, Baid, Wang, Vasudevan, Ku, Yang,
  Ayan, et~al.]{yu2022scaling}
Jiahui Yu, Yuanzhong Xu, Jing~Yu Koh, Thang Luong, Gunjan Baid, Zirui Wang,
  Vijay Vasudevan, Alexander Ku, Yinfei Yang, Burcu~Karagol Ayan, et~al.
\newblock Scaling autoregressive models for content-rich text-to-image
  generation.
\newblock \emph{arXiv preprint arXiv:2206.10789}, 2022.

\bibitem[Yu et~al.(2023)Yu, Sohn, Kim, and Shin]{yu2023video}
Sihyun Yu, Kihyuk Sohn, Subin Kim, and Jinwoo Shin.
\newblock Video probabilistic diffusion models in projected latent space.
\newblock \emph{arXiv preprint arXiv:2302.07685}, 2023.

\bibitem[Zhang et~al.(2021)Zhang, Koh, Baldridge, Lee, and
  Yang]{zhang2021cross}
Han Zhang, Jing~Yu Koh, Jason Baldridge, Honglak Lee, and Yinfei Yang.
\newblock Cross-modal contrastive learning for text-to-image generation.
\newblock In \emph{CVPR}, pages 833--842, 2021.

\bibitem[Zhang and Agrawala(2023)]{zhang2023adding}
Lvmin Zhang and Maneesh Agrawala.
\newblock Adding conditional control to text-to-image diffusion models.
\newblock \emph{arXiv preprint arXiv:2302.05543}, 2023.

\bibitem[Zhou et~al.(2021)Zhou, Wang, Ye, and Zhan]{zhou2021pycil}
Da-Wei Zhou, Fu-Yun Wang, Han-Jia Ye, and De-Chuan Zhan.
\newblock Pycil: A python toolbox for class-incremental learning.
\newblock \emph{arXiv preprint arXiv:2112.12533}, 2021.

\bibitem[Zhou et~al.(2022)Zhou, Wang, Yan, Lv, Zhu, and
  Feng]{zhou2022magicvideo}
Daquan Zhou, Weimin Wang, Hanshu Yan, Weiwei Lv, Yizhe Zhu, and Jiashi Feng.
\newblock Magicvideo: Efficient video generation with latent diffusion models.
\newblock \emph{arXiv preprint arXiv:2211.11018}, 2022.

\bibitem[Zhu et~al.(2019)Zhu, Pan, Chen, and Yang]{zhu2019dm}
Minfeng Zhu, Pingbo Pan, Wei Chen, and Yi~Yang.
\newblock Dm-gan: Dynamic memory generative adversarial networks for
  text-to-image synthesis.
\newblock In \emph{CVPR}, pages 5802--5810, 2019.

\end{thebibliography}



\newpage
\setcounter{section}{0}
\renewcommand{\thesection}{\Roman{section}}
\begin{center}
	\textbf{\large Supplementary of Gen-L-Video }
\end{center}

\makeatletter
\section{Dataset Details}\label{sec:dataset details}
To evaluate our method, we have selected several videos (totaling 66) from TGVE competition~\cite{wu2022tune} and Internet. For internet videos, we have designed 4 distinct prompts that introduce dynamic changes in the areas of object recognition, stylistic elements, background variations, similar motion patterns, or a combination thereof, based on the original prompts.

\begin{table}[h]
\centering
\caption{\textit{\textbf{Names of videos selected.}}}
\label{tab:dataset}
\begin{tabular}{|c|c|c|}
\hline
cows-grazing &  shopping-entertainment-center &  hike \\
warsaw-multimedia-fountain &  deer-eating-leaves &  seagull-flying \\
singapore-airbus-a380-landing &  pigs &  mallard-water \\
red-roses-sunny-day &  cows &  miami-surf \\
sharks-swimming &  pouring-beer-from-bottle &  dog \\
bird-on-feeder &  airbrush-painting &  bear \\
wind-turbines-at-dusk &  horsejump-low &  camel \\
lotus &  typewriter-super-slow-motion &  wetsuit-surfing \\
setting-sun &  cat-in-the-sun &  street-artist-painting \\
gold-fish &  motorbike &  rabbit-watermelon \\
earth-full-view &  snowboard &  eating-pizza \\
fireworks-display &  weightlifting-sofa &  drift-turn \\
surfer-on-wave &  american-flag-in-wind &  basketball-mall \\
ship-sailing &  aircraft-landing &  dirt-road-driving \\
audi-snow-trail &  eiffel-flyover &  butterfly-feeding-slow-motion \\
kettleball-training &  ferris-wheel-timelapse &  dj-mixing-music \\
ski-lift-time-lapse &  swimmer &  raindrops \\
tennis &  swans &  sunset-beach-yoga \\
sunset-swinging &  mbike-trick &  man-skiing \\
car-turn &  boxer-punching-towards-camera &  man-surfing \\
horizontal-match-striking &  airplane-and-contrail &  ski-follow \\
las-vegas-time-lapse &  geometric-video-background &  blackswan \\
\hline
\end{tabular}
\end{table}

\section{User Study Details}\label{sec:user study details}

We utilized the aforementioned dataset as the benchmark to contrast our approach, Gen-L-Video, with the Isolated method. Beyond quantitative indices (frame consistency and textual alignment), we also enlisted several participants to vote on which method was superior. To assess frame consistency, we generated 264 videos from 4 prompts and an additional 65 videos using both methods, respectively. Then we replicate and shuffle them to 1040 videos. We distributed pairs of these videos to several individuals, posing the question, "Which video exhibits superior frame consistency between these two?" In order to gauge textual alignment, we created 279 videos from varying prompts using both methods separately. We subsequently asked the participants, "Which video aligns more accurately with the text description among these two videos? A?" No B nr diffeerenc?? Most participants reported no significant difference in the degree of textual alignment, but a clear improvement in alignment stability, indicating that our method maintains good text-based editing capabilities.The more detailed results of these comparisons are depicted in Table~\ref{tab:quant}.

\section{Proof for the Optimal Approximation}\label{sec:proof}
As we claimed in Sec.~\ref{sec:temporal co-denoising}, the ideal $\vv_{t-1}$ should satisfy that $F_{i}(\vv_{t-1})$ is as close as $\vv_{t-1}^i$ as possible. The optimal $\vv_{t-1}$ can be obtained by solving the following quadratic optimization problem:
\begin{align*}
    \vv_{t-1} = \mathop{\arg\min}_{\vv}\sum_{i=0}^{N-1}\left\|W_i \otimes (F_i(\vv)-\vv_{t-1}^i)\ \right\|_2^2,
\end{align*}
where $W_i$ is the pixel-wise weight for the video clip $\vv^i_{t}$, and $\otimes$ means the tensor product. Here we show that, for an arbitrary frame $j$ in the video $\vv_{t-1}$, namely $\vv_{t-1,j}$, it should be equal to the weighted sum of all the corresponding frames in short videos that contain the $j$ frame.

\begin{proof}
Considering that there are $N$ video clips in total, we denote the set of all video clips and the set of their indices as $\gV=\{\vv^{0},\vv^{1}, \dots, \vv^{N-1}\}$ and $\gI = \{0,1,\dots,N-1\}$, respectively. Further, we assume that there exists a set of video clips $\gV^j$ consisting of short video clips containing the corresponding $j^{th}$ frame $\vv_{\Diamond,j}$ in the original long video. Similarly, the set of indices corresponding to $\gV^{j}$ is denoted as $\gI^j$. Here we use the $\Diamond$ to represent that it could be suitable for all the time $t$ in the denoising path.  

Given a short video clip $\vv^i \in \gV^j$, we denote the corresponding frame of $\vv_{\Diamond,j}$ in $\vv^i$  as $\vv^i_{\Diamond,j^*}$  for simplicity. Note that the $j^*$ in different video clips represents different values. 

Therefore, the original optimization objective can be written as:
\begin{align*}
     & \sum_{i=0}^{N-1}\left\|W_i \otimes \left(F_i(\vv_{t-1})-\vv_{t-1}^i\right)\ \right\|_2^2 \\
    = & 
    \sum_{i\in \gI^j}\left\|W_i \otimes \left(F_i(\vv_{t-1})-\vv_{t-1}^i\right)\ \right\|_2^2  + \sum_{i\in \gI \setminus \gI^j}\left\|W_i \otimes \left(F_i(\vv_{t-1})-\vv_{t-1}^i\right)\ \right\|_2^2 \\
     = & 
    \sum_{i\in \gI^j}\left\|W_{i,j^*} \otimes \left(F_{i,j^*}(\vv_{t-1})-\vv_{t-1,j^*}^i\right)\ \right\|_2^2  + \sum_{i\in \gI^j}\sum_{j\not=j^*}\left\|W_{i,j} \otimes \left(F_{i,j}(\vv_{t-1})-\vv_{t-1,j}^i\right)\ \right\|_2^2 \\   & \qquad + \sum_{i\in \gI \setminus \gI^j}\left\|W_i \otimes (F_i(\vv_{t-1})-\vv_{t-1}^i)\ \right\|_2^2 .
\end{align*} 
Where, $W_{i,j}$ is the pixel-wise weight for the $j^{th}$ frame in $\vv^i_{t-1}$~(\ie $\vv_{t-1,j}^i$), and $F_{i,j}(\vv_{t-1})$ is the $j^{th}$ frame in $F_{i}(\vv_{t-1})$. It is not difficult to observe that the last two terms in the formula have nothing to do with $\vv_{t-1,j}$.  We denote them as constant $C$. Then we have,
\begin{align*}
     & \sum_{i\in \gI^j}\left\|W_{i,j^*} \otimes \left(F_{i,j^*}(\vv_{t-1})-\vv_{t-1,j^*}^i\right)\ \right\|_2^2  + C \\
     = & \sum_{i\in \gI^j}\left[  W_{i,j^*} \otimes \left(F_{i,j^*}(\vv_{t-1})-\vv_{t-1,j^*}^i \right)\right]^\top \left[  W_{i,j^*} \otimes \left(F_{i,j^*}(\vv_{t-1})-\vv_{t-1,j^*}^i \right)\right]+C  \\
     = & \sum_{i\in \gI^j} \left[\left(W_{i,j^*} \otimes F_{i,j^*}(\vv_{t-1}) \right)^\top \left(W_{i,j^*} \otimes F_{i,j^*}(\vv_{t-1}) \right)  + \left(W_{i,j^*} \otimes \vv_{t-1,j^*}^i\right)^\top \left(W_{i,j^*} \otimes \vv_{t-1,j^*}^i\right) \right.\\ &\qquad \left. -2 \left(W_{i,j^*} \otimes \vv_{t-1,j^*}^i\right)^\top\left(W_{i,j^*} \otimes F_{i,j^*}(\vv_{t-1}) \right)  \right]+ C.
\end{align*}
Denote the above objective as $\gL$ and take the gradient of $\gL$ with respect to $\vv_{t-1,j}$, and then we have
\begin{align*}
\frac{\partial \gL}{\partial \vv_{t-1,j}} = 2 \sum_{i\in \gI^j}\left[  \left(W_{i,j^*} \otimes F_{i,j^*}(\vv_{t-1})\right)-\left(W_{i,j^*} \otimes \vv_{t-1,j^*}^i\right)\right] \otimes W_{i,j^*}\otimes \frac{\partial F_{i,j^*}(\vv_{t-1})}{\partial \vv_{t-1,j}}.
\end{align*}
Note that $F_{i,j^*}(\vv_{t-1}) = \vv_{t-1,j}$, therefore $\frac{\partial F_{i,j^*}(\vv_{t-1})}{\partial \vv_{t-1,j}} = \boldsymbol 1$ and we can replace the  $F_{i,j^*}(\vv_{t-1})$ in the above formula with $\vv_{t-1,j}$. Then, we have 
\begin{align*}
\frac{\partial \gL}{\partial \vv_{t-1,j}} & = 2 \sum_{i\in \gI^j}W_{i,j^*}\otimes\left[  \left(W_{i,j^*} \otimes \vv_{t-1,j}\right)-\left(W_{i,j^*} \otimes \vv_{t-1,j^*}^i\right)\right] \\
& = 2 \left[ \sum_{i\in \gI^j} \left(W_{i,j^*} \right)^2\otimes \vv_{t-1,j}-\sum_{i\in \gI^j}\left(\left(W_{i,j^*}\right)^2 \otimes \vv_{t-1,j^*}^i\right)\right].
\end{align*}
Therefore, set the gradient to be zero, and then we get the optimal $\vv_{t-1,j}$,
\begin{align*}
    \vv_{t-1,j} = \frac{\sum_{i\in \gI^j}\left((W_{i,j^*})^2 \otimes \vv_{t-1,j^*}^i\right)}{\sum_{i\in \gI^j} (W^2_{i,j^*})^2},
\end{align*}
which is the weighted sum of all the corresponding frames in short video clips that contain the $j^{th}$ frame.
\end{proof}

\section{Additional Results}\label{sec:addtional results}
Here, we showcase additional results. 

\begin{figure*}[t!]
    \centering
    \vspace{-2cm}  
    \makebox[\textwidth][c]{\includegraphics[width=1.2\textwidth]{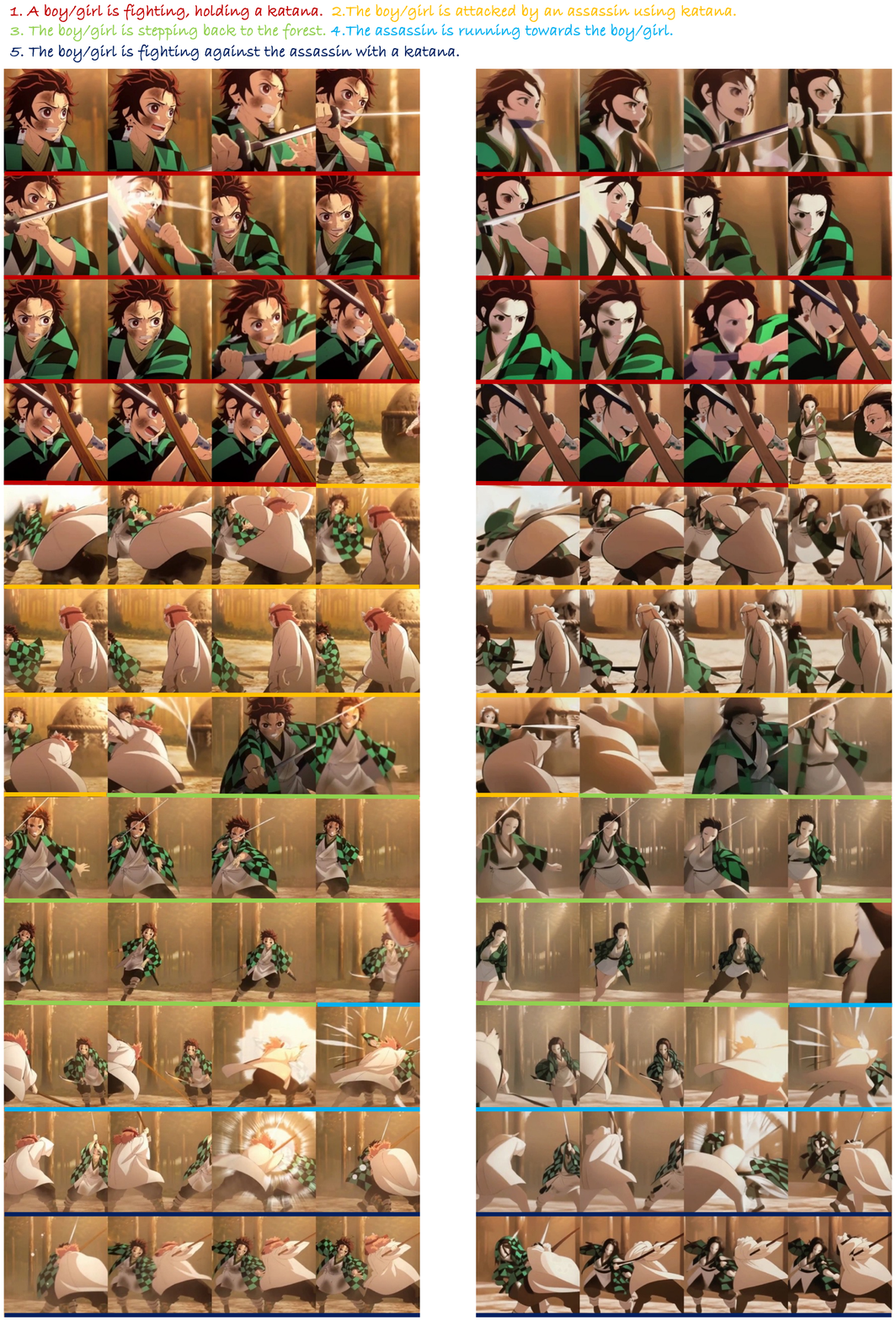}}
    \caption{\textit{\textbf{Multi-text conditioned long video.}}}
    \label{fig:mutli-text}
\end{figure*}

\noindent\textbf{Multi-text long video.} Fig.~\ref{fig:mutli-text} illustrates an example of our multi-text conditioned long video. Specifically, we first split the original video into several short video clips with obvious content changes and of various lengths, and then we label them with different text prompts. The different colors in Fig.~\ref{fig:mutli-text} indicate short videos with different text prompts. Then, we split the original long video into short video clips with fixed lengths and strides. For video clips only containing frames conditioned on the same prompt, we can directly set it as the condition. In contrast, for video clips containing frames conditioned on different prompts, we apply our proposed condition interpolation to get the new condition. After all of these, our paradigm Gen-L-Video can be applied to approximate the denoising path of the long video. 

\begin{figure*}[t!]
    \centering
    \vspace{-2cm}  
    \makebox[\textwidth][c]{\includegraphics[width=1.2\textwidth]{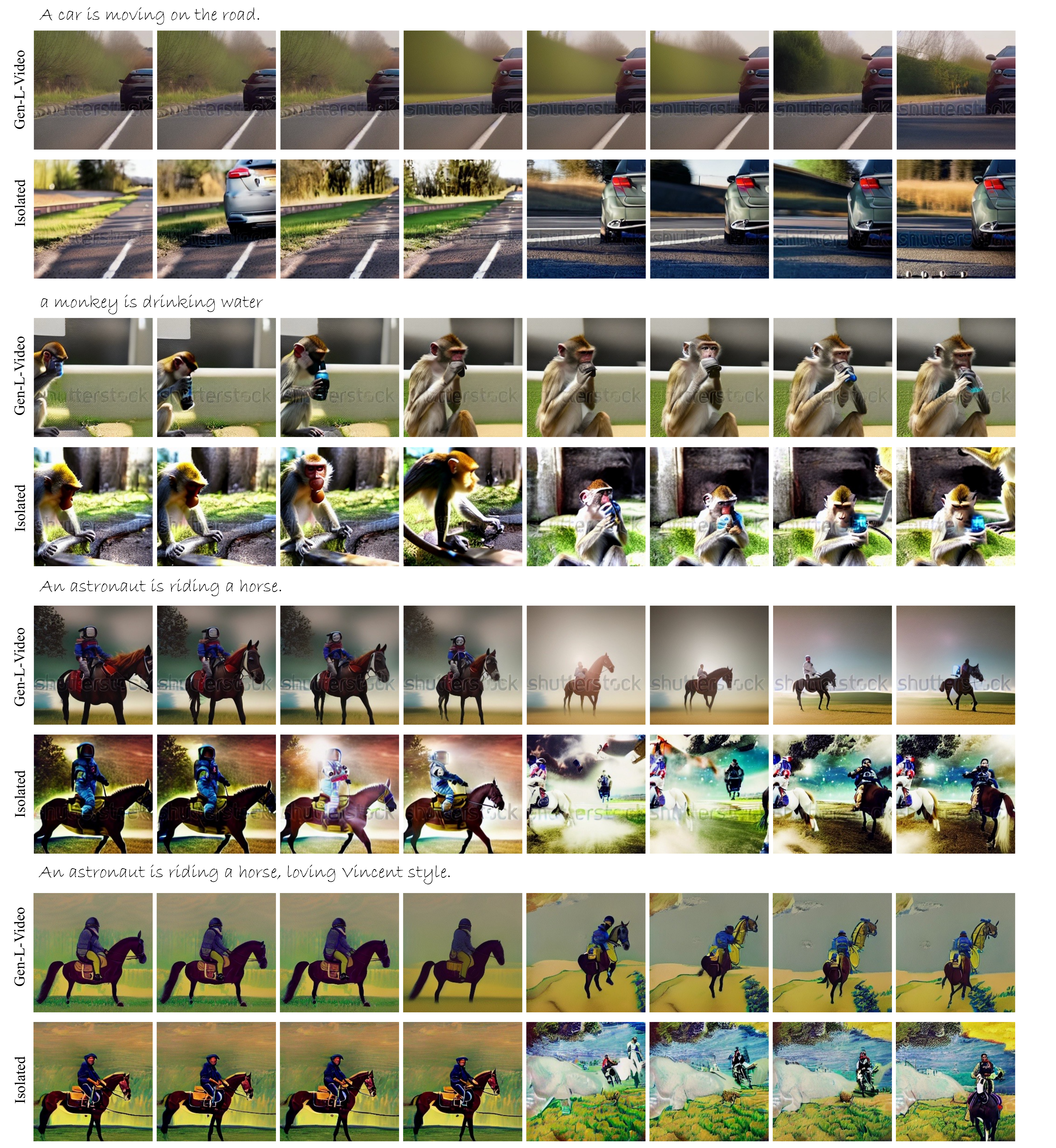}}
    \caption{\textit{\textbf{Long video generation with pretrained short video diffusion models.}}}
    \label{fig:gen}
\end{figure*}

\noindent\textbf{Pretrained Text-to-Video.} Gen-L-Video can also be applied to the pretraiend short video generation model for longer video generation. We compare the results generated through our method and isolated denoising in Fig.~\ref{fig:gen}, and the result reveals that our method significantly enhances the relevance between different video clips.

\begin{figure*}[t!]
    \centering
    \vspace{-2cm}  
    \makebox[\textwidth][c]{\includegraphics[width=1.2\textwidth]{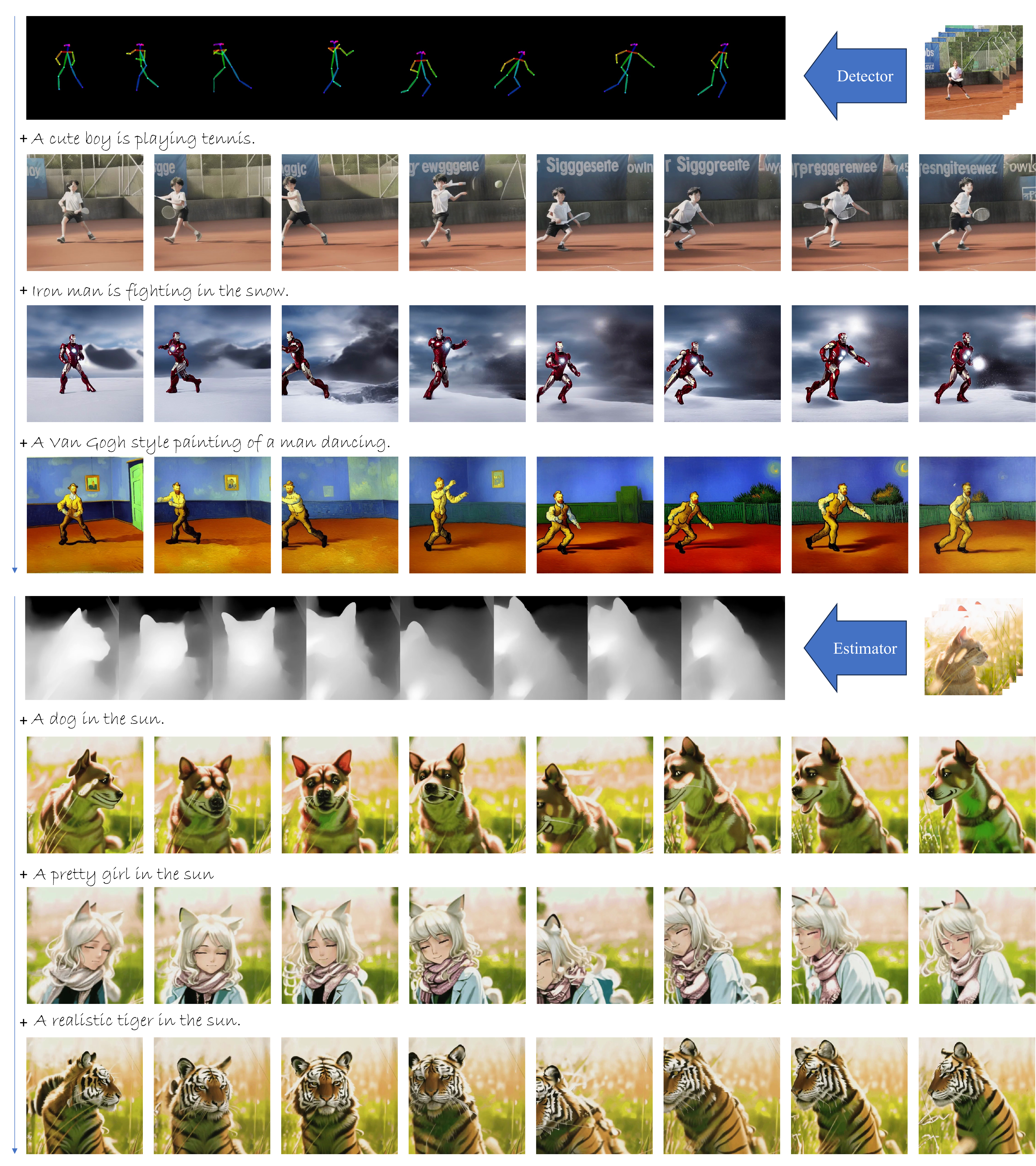}}
    \caption{\textit{\textbf{Controllable long video generation.}}}
    \label{fig:control}
\end{figure*}

\noindent\textbf{Controllabel video generation.} We showcase the results generated by injecting additional control information in Fig.~\ref{fig:control}. The results show that our method can be easily combined with additional information to achieve precise layout control.

\begin{figure*}[t!]
    \centering
    \vspace{-2cm}  
    \makebox[\textwidth][c]{\includegraphics[width=1.2\textwidth]{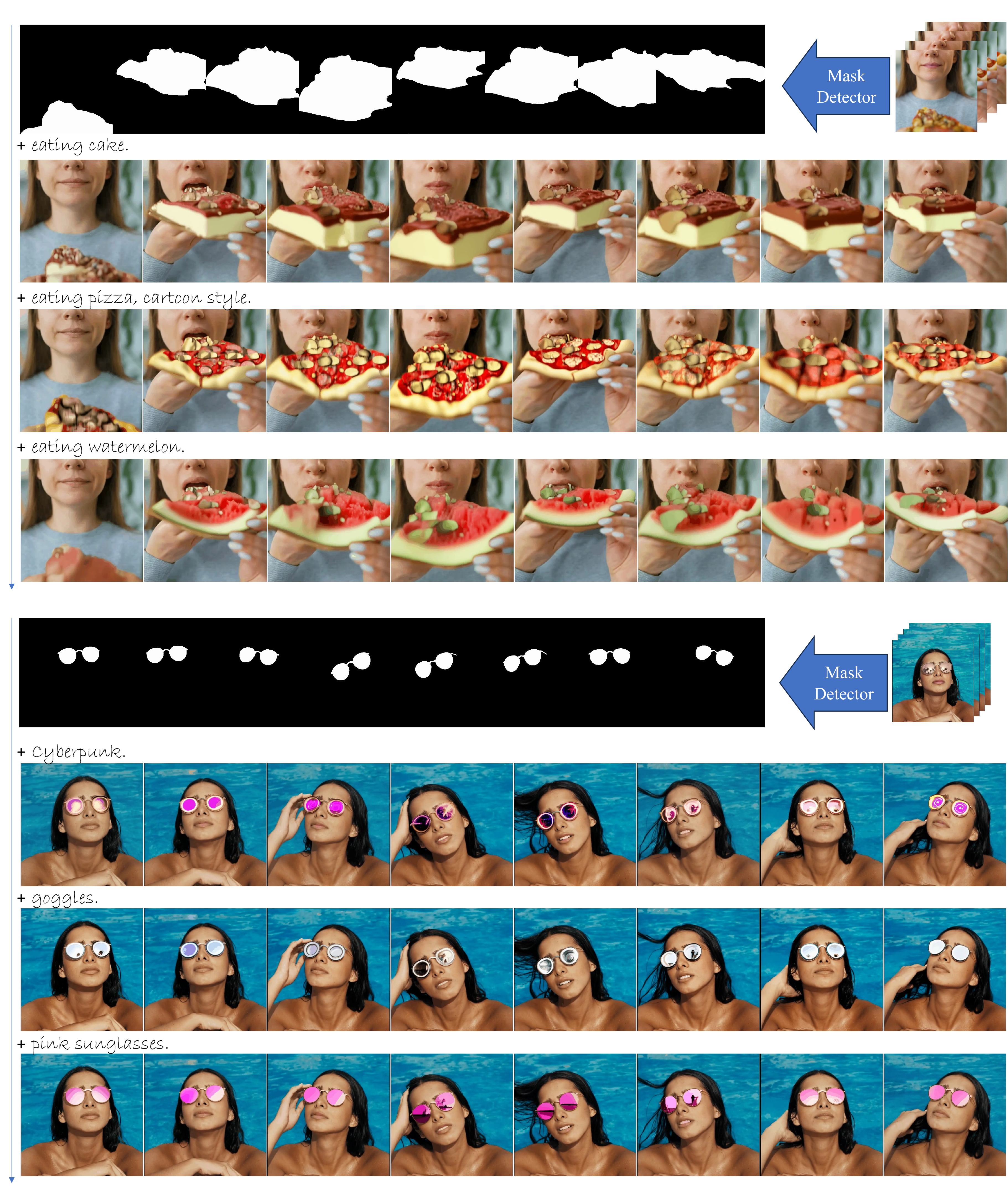}}
    \caption{\textit{\textbf{Edit anything in videos.}}}
    \label{fig:sam_1}
\end{figure*}

\begin{figure*}[t!]
    \centering
    \vspace{-2cm}  
    \makebox[\textwidth][c]{\includegraphics[width=1.15\textwidth]{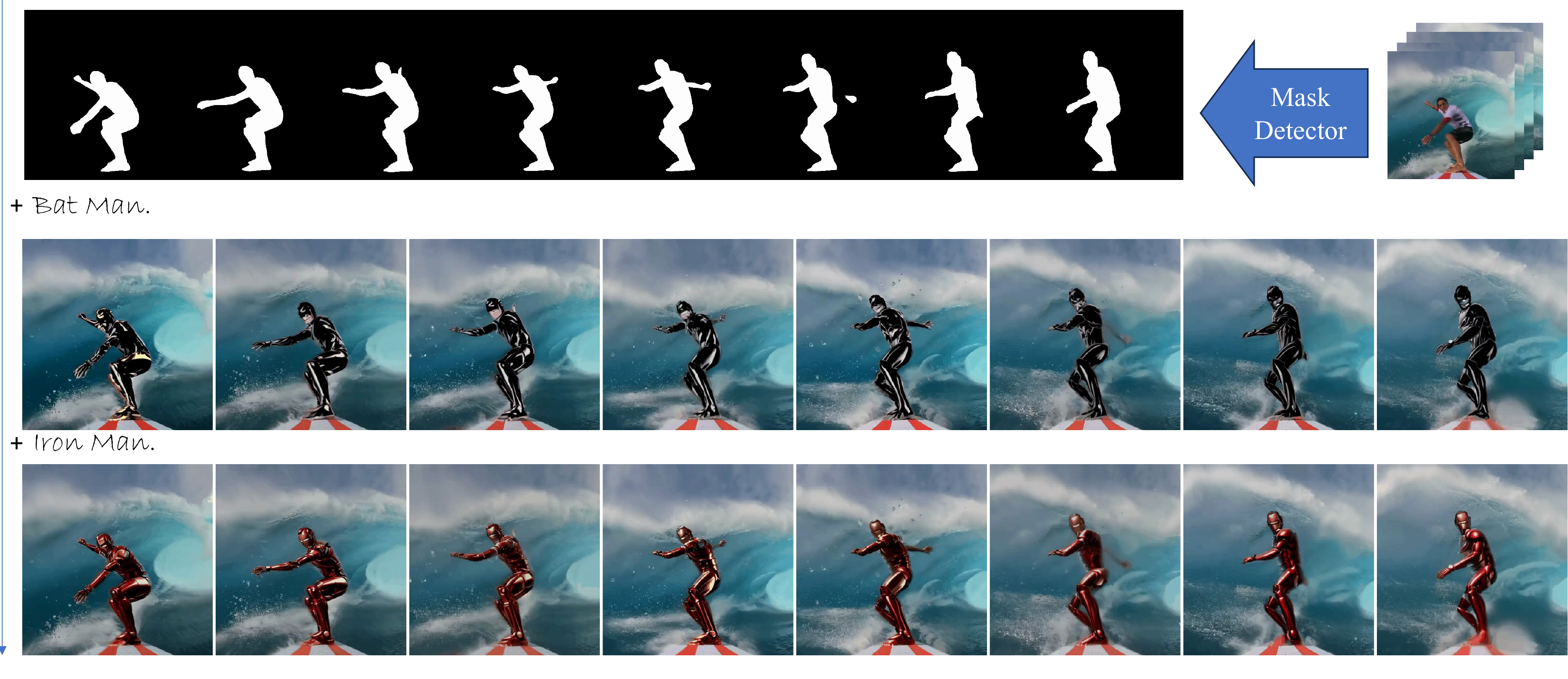}}
    \vspace{-0.5cm}
    \caption{\textit{\textbf{Edit anything in videos.}}}
    \label{fig:sam_2}
\end{figure*}

\noindent\textbf{Edit anything.} Our approach demonstrates significant compatibility with inpainting tasks. As depicted in Fig.\ref{fig:sam_1} and Fig.\ref{fig:sam_2}, our method can reliably edit very long videos and maintain consistent content. The examples given in Fig.~\ref{fig:sam_1} are longer than 12 and 20 seconds, respectively.

\begin{figure*}[t!]
    \centering
    \vspace{-0.5cm}  
    \makebox[\textwidth][c]{\includegraphics[width=1.15\textwidth]{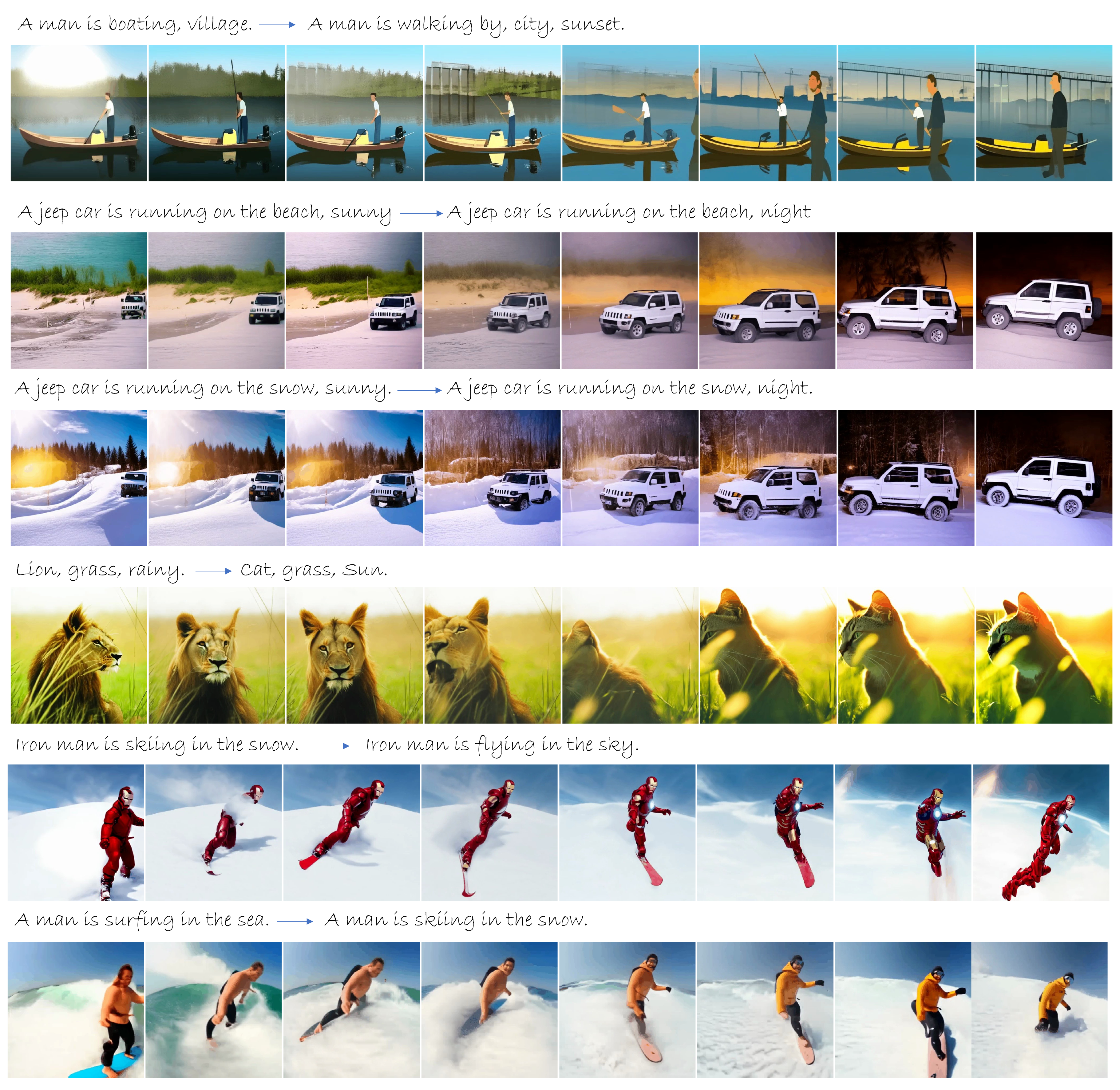}}
    \caption{\textit{\textbf{Long videos with smooth semantic changes.}}}
    \label{fig:semantic-changes}
\end{figure*}

\noindent\textbf{Long video with smooth semantic changes.} Our paradigm also allows us a pleasant application where we are able to edit the source video to generate videos with smooth semantic changes. For example, we are able to generate cars running on the road from day to night to reflect the time flies. The generated results are represented in Fig.~\ref{fig:semantic-changes}.


\end{document}